\pdfoutput=1

\documentclass[11pt]{article}

\usepackage[final]{acl}

\usepackage{times}
\usepackage{latexsym}

\usepackage[T1]{fontenc}

\usepackage[utf8]{inputenc}

\usepackage{microtype}

\usepackage{inconsolata}

\usepackage{graphicx}
\usepackage{booktabs}
\usepackage{subcaption}

\title{Is artificial intelligence still intelligence? LLMs generalize to novel adjective-noun pairs, but don't mimic the full human distribution}

\author{
 \textbf{Hayley Ross\textsuperscript{1}},
 \textbf{Kathryn Davidson\textsuperscript{1}},
 \textbf{Najoung Kim\textsuperscript{2}}
\\
 \textsuperscript{1}Harvard University
 \textsuperscript{2}Boston University
\\
 \small{
   \texttt{hayleyross@g.harvard.edu} \quad 
   \texttt{kathryndavidson@fas.harvard.edu} \quad
   \texttt{najoung@bu.edu}
 }
}

\begin{document}
\maketitle
\begin{abstract}
Inferences from adjective-noun combinations like \textit{Is artificial intelligence still intelligence?} provide a good test bed for LLMs' understanding of meaning and compositional generalization capability, since there are many combinations which are novel to both humans and LLMs but nevertheless elicit convergent human judgments. 
We study a range of LLMs and find that 
the largest models we tested are able to draw human-like inferences when the inference is determined by context and can generalize to unseen adjective-noun combinations. We also propose three methods to evaluate LLMs on these inferences out of context, where there is a distribution of human-like answers rather than a single correct answer. We find that LLMs show a human-like distribution on at most 75\% of our dataset, which is promising but still leaves room for improvement.
\end{abstract}

\section{Introduction}

As the pretraining datasets of LLMs grow, it becomes increasingly difficult to test whether LLMs can generalize to unseen instances of linguistic phenomena, since it is hard to systematically exclude them from the pretraining data \citep{kim_uncontrolled_2022}. 
Adjective-noun combinations, in particular with so-called privative adjectives like \textit{fake}, provide a good test bed for testing granular language understanding and semantic composition in LLMs, since we can find many adjective-noun bigrams which are easy for humans to understand despite being novel to them, and can further confirm that they are novel to the LLM by a string search over its pretraining corpus.
Adjective-noun bigrams are naturally associated with membership inferences which allow us to test whether LLMs have successfully ``composed'' their meaning: is an \{adjective\} \{noun\} still a \{noun\}? This inference is not as straightforward as it may seem: while a \textit{yellow flower} is clearly still a \textit{flower} (a \textit{subsective} inference), a \textit{fake} or \textit{counterfeit dollar bill} is typically not a \textit{dollar bill} (a \textit{privative} inference), even though a \textit{fake watch} is typically a \textit{watch} \citep{martin_compositional_2022,ross_fake_2024}.
In order to draw the correct inference, humans and LLMs need to grasp not only the effect of the adjective on the noun's properties, but also which properties are required vs.~merely typical for membership in that noun/category. 
Moreover, like many cases of meaning, this inference depends on context. For example, a \textit{fake crowd} might qualify as a \textit{crowd} if it is made up of paid actors, but less so if it is just painted dummies on a movie set.
Nonetheless, humans derive convergent inferences for many novel bigrams both with and without context, giving a ground truth to which we can compare LLMs.

\begin{figure}[t]
  \includegraphics[width=\columnwidth]{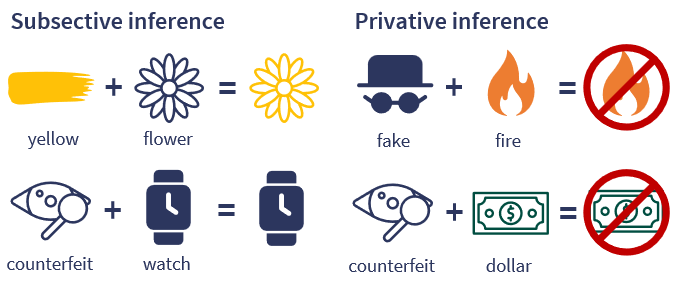}
  \caption{Membership inferences for adjective-noun combinations vary by adjective and noun.}
  \label{fig:inference-examples}
\end{figure}

Building on the dataset of English human judgments developed in \citet{ross_fake_2024}, we compare LLMs of several sizes with humans for cases in which the context determines the inference. We further explore three methods to evaluate LLMs against the full distribution of human ratings when there is no provided context.
We find that when the context determines the inference, recent larger LLMs closely match human behavior, 
while smaller LLMs only sometimes do so. 
Almost all LLMs, even smaller ones, are able to handle novel adjective-noun combinations as well as they handle known ones, suggesting that they evaluate these combinations on-the-fly and can generalize accordingly like humans do. This alignment in performance on observed and novel combinations continues to hold in a setting where no context is provided. 
However, we find that even 70B parameter LLMs 
are only able to approximate the distribution of ratings of a population of human raters for 75\% of our dataset or less, struggling on combinations with high human variance and inferences which are rare for a particular adjective, such as a \textit{homemade cat} not necessarily being a \textit{cat}.
Most LLMs also assign a more positive rating to the question ``Is artificial intelligence still intelligence?'' than most humans (see Figure \ref{fig:intelligence-ratings}).
In sum, while our generalization results are highly promising, there is room for improvement on the task of matching human inferences in this category overall. We share our code and data on GitHub.\footnote{\url{https://github.com/rossh2/artificial-intelligence/}}

\section{Related work}

Most previous computational work on adjective-noun composition focuses on distributional semantics using word embeddings \citep{boleda_intensionality_2013,boleda_first_2012,cappelle_facing_2018,guevara_regression_2010,hartung_learning_2017,vecchi_spicy_2017}. 
\citet{cappelle_facing_2018} specifically analyze privative adjective embeddings, finding no difference between their embeddings and embeddings of other adjectives. 
Results for early LLMs are largely negative:
BERT \citep{devlin_bert_2019} shows no evidence of compositionality for adjective-noun and noun-noun phrases, relying instead on word overlap heuristics for similarity judgments \citep{yu_assessing_2020}.
\citet{bertolini_testing_2022} study the inferences of adjective-noun combinations for BERT and RoBERTa \citep{liu_roberta_2019}. They divide adjectives into three inference classes---intersective, subsective and privative---based on previous computational work \citep{lalisse_distinguishing_2015,nayak_dictionary_2014}, and test whether LLMs can draw the correct pattern of inferences for adjectives in each class. 
However, recent work in linguistics \citep{pavlick_so-called_2016,martin_compositional_2022,ross_fake_2024} suggests that \citeauthor{bertolini_testing_2022}'s task may be ill-defined, since adjectives in these ``classes'' can license either a subsective or a privative inference depending on the noun and context. 
Indeed, \citet{goodale_systematic_2023} find that BERT can distinguish between intersective and subsective adjectives, which are more clearly distinct.

For the ``Is an X a Y'' task more broadly, 
LLMs from BERT to Llama 2 \citep{touvron_llama_2023-1} show some, but limited abilities to judge ``Is an X a Y'' for hypernyms, 
especially with negation \citep{hanna_analyzing_2021,ettinger_what_2020,ravichander_systematicity_2020,nikishina_predicting_2023,moskvoretskii_are_2024}. 
Results from property learning also show that earlier models struggle to learn what properties are typical of nouns \citep{do_are_2021, apidianaki_all_2021,pavlick_most_2016}---a key part of our task, since LLMs must know what constitutes a \textit{watch} in order to judge whether a \textit{fake watch} counts as one. 
Meanwhile, 
\citet{lyu_is_2022} find that BERT and GPT-3 \citep{brown_language_2020} cannot handle inferences on recursive adjectives, such as ``Is my favorite new movie my favorite movie?'', while \citet{sathe_language_2024} find that GPT-2 \citep{radford_language_2019}, MPT 30B \citep{MosaicML2023Introducing} and other smaller models struggle to predict whether rare adjective-noun combinations are acceptable to humans.
However, less is known about the capabilities of newer, larger models in these areas.
Recent work with newer models including Llama 2 and GPT-3 on noun-noun compounds---whose meaning arises less straightforwardly from their parts than adjective-noun combination \citep{hacken_semantics_2016}---shows that the LLMs do not generalize well in this case 
\citep{ormerod_how_2024,coil_chocolate_2023,rambelli_can_2024}.

\section{Human judgment dataset} 

\citet{ross_fake_2024} present two datasets of human judgments on adjective-noun inferences. The first, which we refer to as \textsc{No-Context}, collects human ratings on ``Is an \{adjective\} \{noun\} a \{noun\}?'' on a 5-point Likert scale (``Definitely not'', ``Probably not'', ``Unsure'', ``Probably yes'' and ``Definitely yes'') for 798 bigrams, covering 102 nouns crossed with 6 typically-privative and 6 typically-subsective adjectives. 
(In this paper, we use ``(typically-)privative / subsective adjective'' to refer to adjectives historically classed as such, which often but not always result in the respective inference.)
180 of the 798 bigrams are zero frequency in the C4 pretraining corpus \citep{raffel_exploring_2020}, which we take as a proxy for the undisclosed pretraining corpora of the models we study. We take these bigrams to be novel to both humans and LLMs.\footnote{The dataset was not published when these models were trained, so there is no danger of it being included in model pretraining.} 
We call a bigram high-frequency if it is in the top quartile of bigrams studied by \citeauthor{ross_fake_2024} 

\citeauthor{ross_fake_2024} show that this inference depends on the adjective and noun, with bigrams with ``subsective'' adjectives usually (but not always; e.g., \textit{homemade cat}) being rated subsective, while bigrams with ``privative'' adjectives such as \textit{fake crowd} elicit a wide distribution of ratings from privative to subsective, with high variance per bigram. 
Moreover, humans converge on inference judgments for many zero-frequency bigrams, and show similar variance overall between zero-frequency and high-frequency bigrams, demonstrating that they can generalize these inferences.

The second dataset, which we refer to as \textsc{Context}, shows that providing a context with appropriate detail is sufficient to determine the inference for typically-privative adjectives. 
Participants first read a short 50-word context and then answer ``In this setting, is an \{adjective\} \{noun\} a \{noun\}?''
The dataset contains 56 expert-written contexts for 28 bigrams, with one privative-biased and one subsective-biased context for each bigram; an example is shown in Appendix \ref{app:context-examples}. 
6 bigrams are zero-frequency in C4 and a further 7 are low-frequency (below median frequency in the total set). This dataset is much smaller due to the need for trained annotators to create the contexts.

\section{Experiment 1: In-context meaning generalization} \label{sec:context-experiment}

Typical LLM evaluations assume a single correct answer for each question. Thus, we begin with the smaller \textsc{Context} dataset from \citet{ross_fake_2024} where the contexts provided are sufficient to determine the inference. 
Our evaluation focuses on the following two aspects: (1) whether the LLM is sensitive to the provided context, and (2) whether the LLM is able to choose the intended inference both for high-frequency bigrams it has presumably seen during pretraining and also for zero-frequency bigrams which we presume it has not. 

\subsection{Method}

Experiment 1a evaluates the 28 bigrams in \textsc{Context} with no context provided.
While there is no single correct rating in this setting, we can use this as a baseline to see if providing a context changes the rating.
Experiment 1b evaluates the same bigrams but provides the two contexts for each bigram which bias the rating for humans. 
For both, we adapt the method used in \citet{ross_fake_2024} as closely as possible for LLMs. We use the same question wording, asking ``(In this context [1b],) is an \{adjective\} \{noun\} still a \{noun\}?'' 
with the same 5-point Likert scale, the only difference being that the Likert scale is described in words rather than pictured. We provide 5 few-shot examples illustrating each rating on the scale.\footnote{Experiment 1c in Appendix \ref{app:zero-shot} performs an ablation study on Experiment 1b with 0-shot prompting.} 
We believe this to be a comparable setup since humans also see examples during the training phase of the experiment, though humans only see three.
The few-shot examples only demonstrate the use of the Likert scale with ``is-a'' judgments, and do not include any typically-privative adjectives; see Appendix \ref{app:few-shot-examples}.

To get responses on the Likert scale, we calculate the surprisal
of the 5 answers.
The model's response is whichever of these 5 answers has the lowest surprisal. This limits us to assessing open-source models where log-probabilities of the input are available, excluding popular closed-source models like GPT-4 \citep{openai_gpt-4_2024}. We study the Llama 2 \citep{touvron_llama_2023-1} and Llama 3 series\footnote{In this paper, Llama 3 refers to the original Llama 3 models, not the newly released Llama 3.1 models.} \citep{dubey_llama_2024} in detail, as well as Mixtral 7x8B \citep{jiang_mixtral_2024} and Qwen 2 72B \citep{yang_qwen2_2024}.
We test all sizes of Llama 2 and Llama 3 to investigate whether generalization ability improves with model size. We primarily focus on instruction-tuned models; results for the base Llama models are shown in Appendix \ref{app:result-figures-tables}.

\subsection{Results}

\begin{figure*}[t]
  \includegraphics[width=\textwidth]{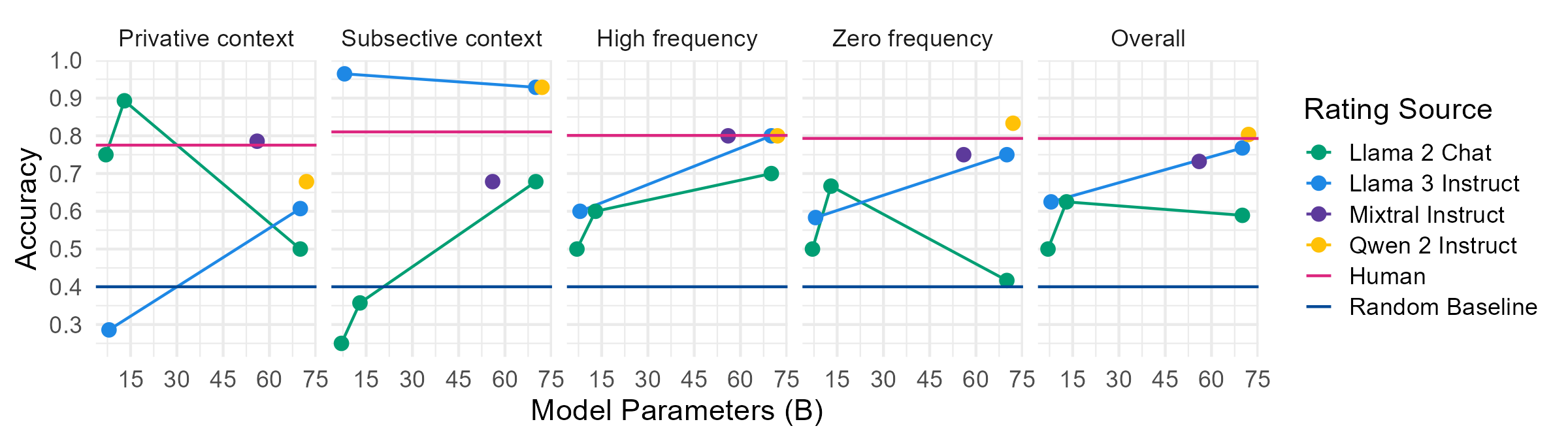}
  \caption{Accuracy on the context-based inference task (Experiment 1b) overall, in privative vs.~subsective contexts, and for high frequency vs.~zero frequency bigrams.
  Accuracy on the context-based inference task increases with model parameters for all models except Llama 2 Chat, and all models except Llama 2 70B Chat can generalize to (perform similarly or better on) zero frequency (novel) bigrams.
  }
  \label{fig:scaling}
\end{figure*}

Firstly, to measure whether the provided biased contexts have a significant effect on inference ratings, we compare the results from Experiment 1a and 1b. 
Detailed results for Experiment 1a are given in Appendix \ref{app:exp1a}. 
We fit the same ordinal regression as \citet{ross_fake_2024}, \verb|Rating ~ ContextBias|, in R \citep{r_core_team_r_2023,christensen_ordinalregression_2022}. 
While \citet{ross_fake_2024} find that for humans, both the privative and subsective contexts have a significant effect on ratings compared to rating the bigram with no context, we find that this is not the case for all LLMs. 
For all LLMs,
we find that subsective contexts have a significant effect ($p < 0.05$) compared to providing no context in Experiment 1a.
Privative contexts only have a significant effect ($p < 0.05$) for 
those models which rate many bigrams with typically-privative adjectives as subsective without context (see Figure \ref{fig:exp1a-no-context} in Appendix \ref{app:exp1a}), namely
Llama 3 70B Instruct, Llama 3 8B Instruct and Qwen 2 72B Instruct. 

Next, 
we judge the inference as correct if the rating is ``Definitely/Probably not'' in privative-biased contexts and ``Probably/Definitely yes'' in subsective-biased contexts.
Figure \ref{fig:scaling} shows the accuracy for all language models under this metric, plus a random guessing baseline.
The human results in Figure \ref{fig:scaling} should be viewed as a ceiling measuring the effectiveness of the context at fixing the inference, not human competence at the task.   

We see that Llama 3 70B Instruct, Qwen 2 72B Instruct and Mixtral 7x8B Instruct perform similarly to humans on this task, suggesting a good ability to (a) understand the effect the context has on the thing described, (b) understand what is necessary to count as an instance of each noun, and (c) draw the correct inference based on the previous two steps. 
For all models except Llama 2 70B Chat, we see no difference between their accuracy on high-frequency bigrams
and zero-frequency ones.

Likewise, performance scales with model size for all models except Llama 2 70B Chat. 
This is because this metric penalizes use of the ``Unsure'' rating, which Llama 2 70B Chat often uses (see Figure \ref{fig:exp1a-no-context}), and uses more often for the 6 zero-frequency bigrams. Llama 2 13B Chat scores higher because it is more confident, even though it is sometimes confidently wrong.
(Under the softer metric of ``accuracy within 1 SD of the human mean'', which we will introduce in Section \ref{sec:acc-1-sd-method}, performance does indeed scale with model parameters -- see Figure \ref{fig:scaling-instruct-1sd} in the Appendix.)

\subsection{Discussion}

While model performance on judging inferences of adjective-noun combinations given a context improves with scale, we see that almost all models behave similarly for high-frequency and zero-frequency bigrams, despite presumably never having seen the zero-frequency bigrams before (or at least despite the substantial frequency gap). 
Thus, we conclude that models do not handle this task by memorizing inferences of noun membership during pretraining and instead assess it dynamically, thus being able to generalize (whether this process is a case of genuine adjective-noun meaning composition, as humans are presumed to be able to do in linguistic theory, or some other heuristics). 
While this is an exciting result,  
this is a rather small dataset. 
In the next section, we study the larger \textsc{No-Context} dataset from \citet{ross_fake_2024}.

\begin{figure*}[t]
  \includegraphics[width=\textwidth]{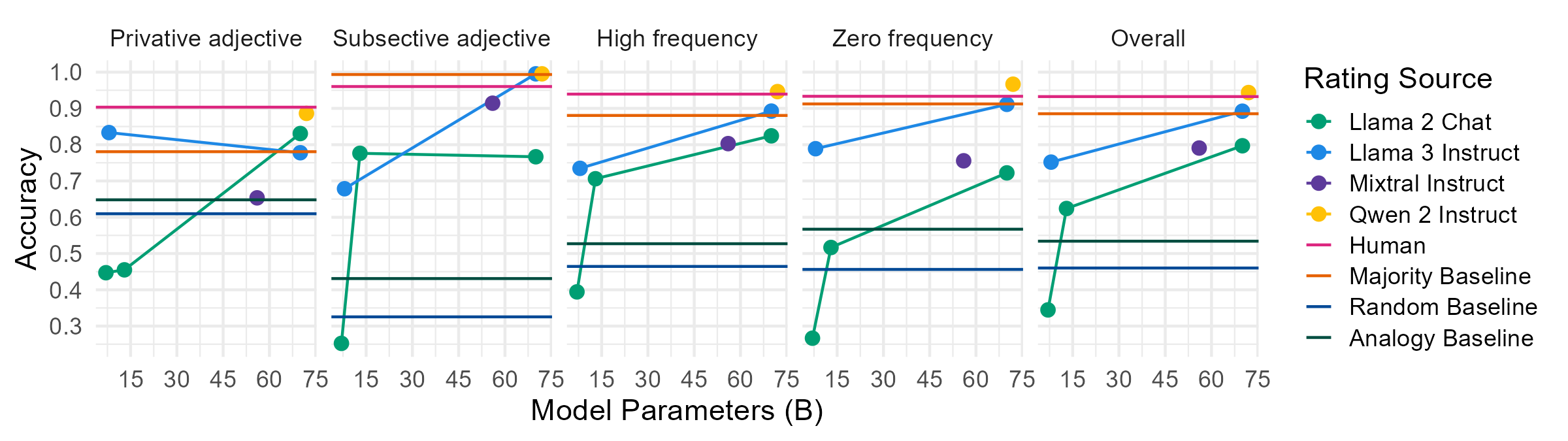}
  \caption{Accuracy within 1 SD of the human mean on the no-context inference task (Experiment 2) overall, for typically privative vs.~subsective adjectives, and for high vs.~zero frequency bigrams. While accuracy is high, a simple ``majority'' baseline nearly saturates this metric.
  }
  \label{fig:contextfree-1sd-scaling}
\end{figure*}

\begin{figure}[t]
  \includegraphics[width=\columnwidth]{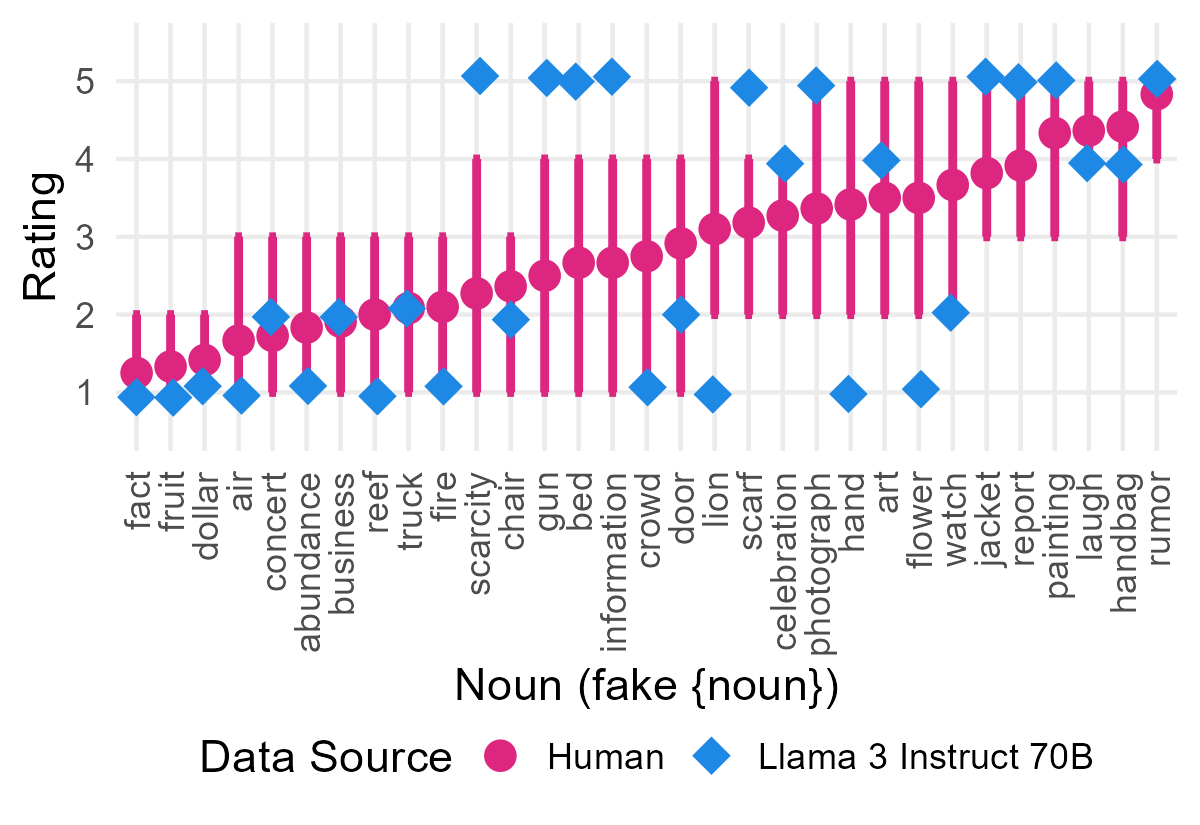}
  \caption{
  Ratings for select bigrams involving \textit{fake} for Llama 3 Instruct 70B, compared to the (rounded) 1 SD interval around the human mean.}
  \label{fig:fake-accuracy-1sd}
\end{figure}

\section{Experiments 2 and 3: Evaluating inferences without context} \label{sec:isa-data-no-context}

The \textsc{No-Context} dataset in \citet{ross_fake_2024} asks the same inference question ``Is an \{adjective\} \{noun\} a \{noun\}?'', but without providing any additional context to help determine the inference. This results in a wide distribution of human ratings for some (but not all) bigrams involving typically privative adjectives. For example, a \textit{counterfeit dollar bill} is never judged to be a \textit{dollar bill}. The distribution also widens for some bigrams with typically subsective adjectives, such as \textit{homemade cat}. 
In this section, we propose three methods to investigate whether LLMs can match the distribution of ratings provided by humans in this context, since there is no longer a single correct answer.

\subsection{Method 1: Accuracy within 1 SD} \label{sec:acc-1-sd-method}

The first method evaluates a single judgment derived from an LLM  
and asks: Is the LLM sampling from the same distribution as the humans for each bigram?
A quick but coarse metric to assess this is whether the LLM's rating falls within one standard deviation of the human mean, rounded to the nearest integer rating (1 corresponds to ``Definitely not'', 5 to ``Definitely yes'').\footnote{One alternative would be to compute the inter-annotator agreement (IAA) between the LLM and the other, human annotators, but existing metrics for IAA either calculate the agreement between two annotators or the agreement across the whole group.}
This method is intuitive and easy to compute, but is also a relatively low bar. To illustrate, Figure \ref{fig:fake-accuracy-1sd} shows the human means with intervals 1 SD wide for a selection of bigrams of the form \textit{fake} \{noun\}, with the no-context ratings from Llama 3 Instruct 70B superimposed. For \textit{fake} in particular, these intervals are relatively wide and easy for the LLM to land in, although for subsective adjectives like \textit{useful}, these intervals are much smaller. 
Conversely, subsective inferences are by far the most common inference, so we still expect high performance for subsective adjectives.

Experiment 2 obtains single ratings for each of the 798 bigrams in \textsc{No-Context} 
using the same setup as Experiment 1a (Section \ref{sec:context-experiment}). 
Figure \ref{fig:contextfree-1sd-scaling} shows the scores 
on this metric, split by adjective type (typically privative or typically subsective) and bigram frequency. 
We also compare to three baselines: random, ``majority'', and analogy. For the random baseline, we sample 100 ratings from 1--5 for each bigram , calculate whether it is within 1 SD of the human mean for that bigram, and average the results.
The ``majority'' baseline guesses a fixed rating depending on the adjective's underlying category as typically subsective or typically privative. Bigrams with subsective adjectives are rated ``Definitely yes'' (5), while bigrams with privative adjectives are rated ``Unsure'' (3), taking advantage of the fact that privative adjectives tend to have wide 1-SD intervals that often overlap with 3. 

Finally, the analogy baseline attempts to calculate the inference by analogy, imitating reasoning such as ``a fake watch is a watch, and a handbag is an expensive accessory like a watch, so a fake handbag must also be a handbag''. Specifically, it uses the distance between GloVe embeddings \citep{pennington_glove_2014} 
to find nearby adjectives and nearby nouns among all the adjectives and nouns used in \citet{ross_fake_2024}, assembling those into ``nearby bigrams''. It then averages the human inference ratings among those nearby bigrams which are high-frequency (assumed ``known'') to predict the rating of the new bigram.

We find that most models, with the exception of Llama 2 7B Chat, perform well under this metric, but the high performance of the majority baseline highlights the leniency of this metric. In fact, only Llama 3 70B Instruct manages to outperform the majority baseline. That said, the gap between the analogy baseline and the larger models suggests that models are doing something more sophisticated than simple analogical reasoning to high-frequency inferences that they have seen before. 
This is further borne out by the models' high performance on zero-frequency bigrams, as in Experiment 1b, showing again that models seem to handle this task on-the-fly rather than relying on having seen the combination during pretraining. As in Experiment 1b, performance scales with size.

\begin{figure}[t]
  \includegraphics[width=\columnwidth]{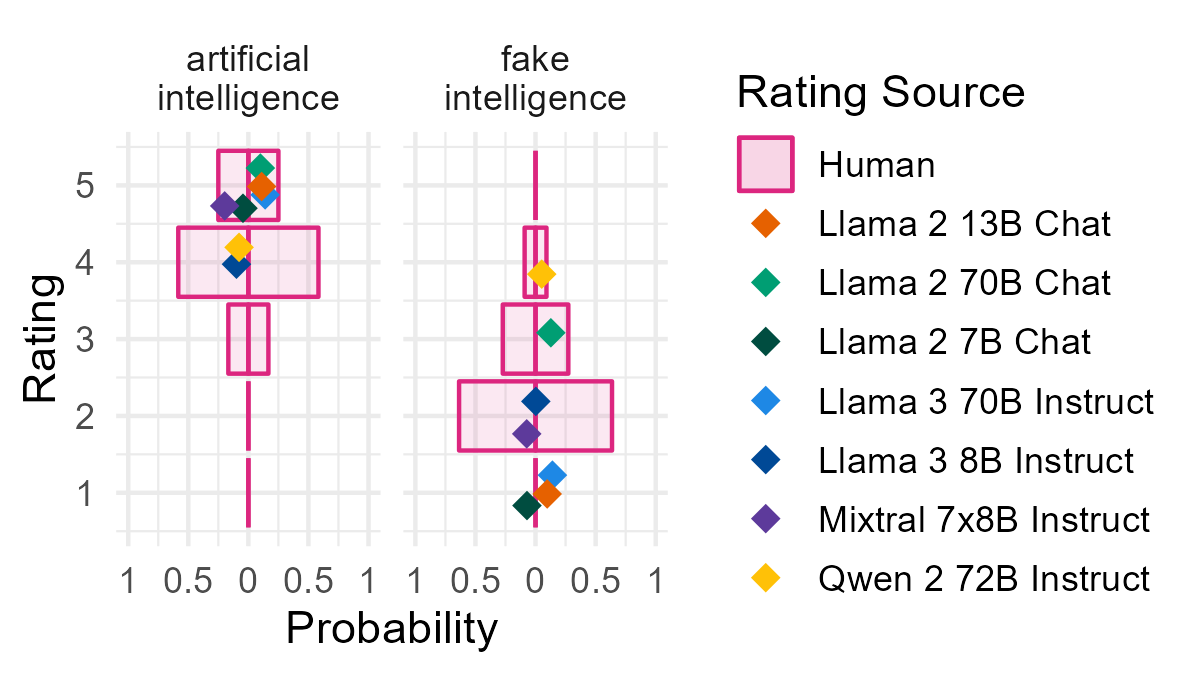}
  \caption{Ratings for ``\textit{Is artificial/fake intelligence still intelligence?}'', showing the distribution for humans and the single rating (with no context provided) for LLMs. Most instruction-tuned LLMs give a more confident (higher) rating than humans for \textit{artificial intelligence}.}
  \label{fig:intelligence-ratings}
\end{figure}

For \textit{artificial intelligence} specifically, we find that most LLMs answer ``Is artificial intelligence still intelligence?'' with ``Definitely yes'', while humans tend to prefer a more conservative ``Probably yes'', as shown in Figure \ref{fig:intelligence-ratings}.

While this method is appealingly simple,  
performance on this metric is close to saturated by the majority baseline, making it difficult
to evaluate whether LLMs are performing in a ``human-like'' way. 
We next explore two methods of generating a distribution of ratings from an LLM, to see if  LLMs can capture the whole human distribution 
rather than merely capturing a point within it.

\subsection{Method 2: Log-probability distribution} \label{sec:sampling-logprobs}

\subsubsection{Method}

Our second method obtains a distribution of ratings from the LLM by calculating the log-probabilities of all 5 answers for each of the 798 bigrams in \textsc{No-Context} in Experiment 2 and 
converting this into a probability distribution for each bigram. 
For each bigram, we calculate the Jensen-Shannon divergence between the distribution of ratings obtained from the LLM and the distribution given by the (normalized) human ratings.\footnote{While calculating the Kullback-Leibler divergence would also let us treat the human ratings as ground truth, we prefer the Jensen-Shannon divergence because it is bounded between 0 and 1 and thus easier to interpret.} 
0 indicates perfect overlap, while 1 indicates maximal divergence. 

\subsubsection{Results}

\begin{table}[t]
    \centering
    \begin{tabular}{lccc}
    \toprule
    & \multicolumn{3}{c}{JS Divergence} \\
    Model & Priv. & Subs. & Total \\ \midrule
    Human & 0 & 0 & 0 \\
    Llama 3 70B Instruct & 0.26 & 0.08 & 0.17 \\
    Qwen 2 72B Instruct & 0.33 & 0.08 & 0.19  \\
    Llama 2 70B Chat & 0.18 & 0.25 & 0.22 \\
    Mixtral 7x8B Instruct & 0.32 & 0.13 & 0.22 \\
    Llama 3 8B Instruct & 0.18 & 0.34 & 0.26 \\
    Llama 2 13B Chat  & 0.25 & 0.35 & 0.30 \\
    Uniform baseline & 0.20 & 0.46 & 0.34 \\
    Llama 2 7B Chat  & 0.29 & 0.46 & 0.38 \\ 
    ``Majority'' baseline & 0.71 & 0.12 & 0.40 \\
    \bottomrule
    \end{tabular}
    \caption{Jensen-Shannon divergence between per-bigram rating distributions for humans and LLM log-probabilities, for privative vs.~subsective adjectives.}
    \label{tab:logprob-js}
\end{table}

\begin{figure}[t]
  \includegraphics[width=\columnwidth]{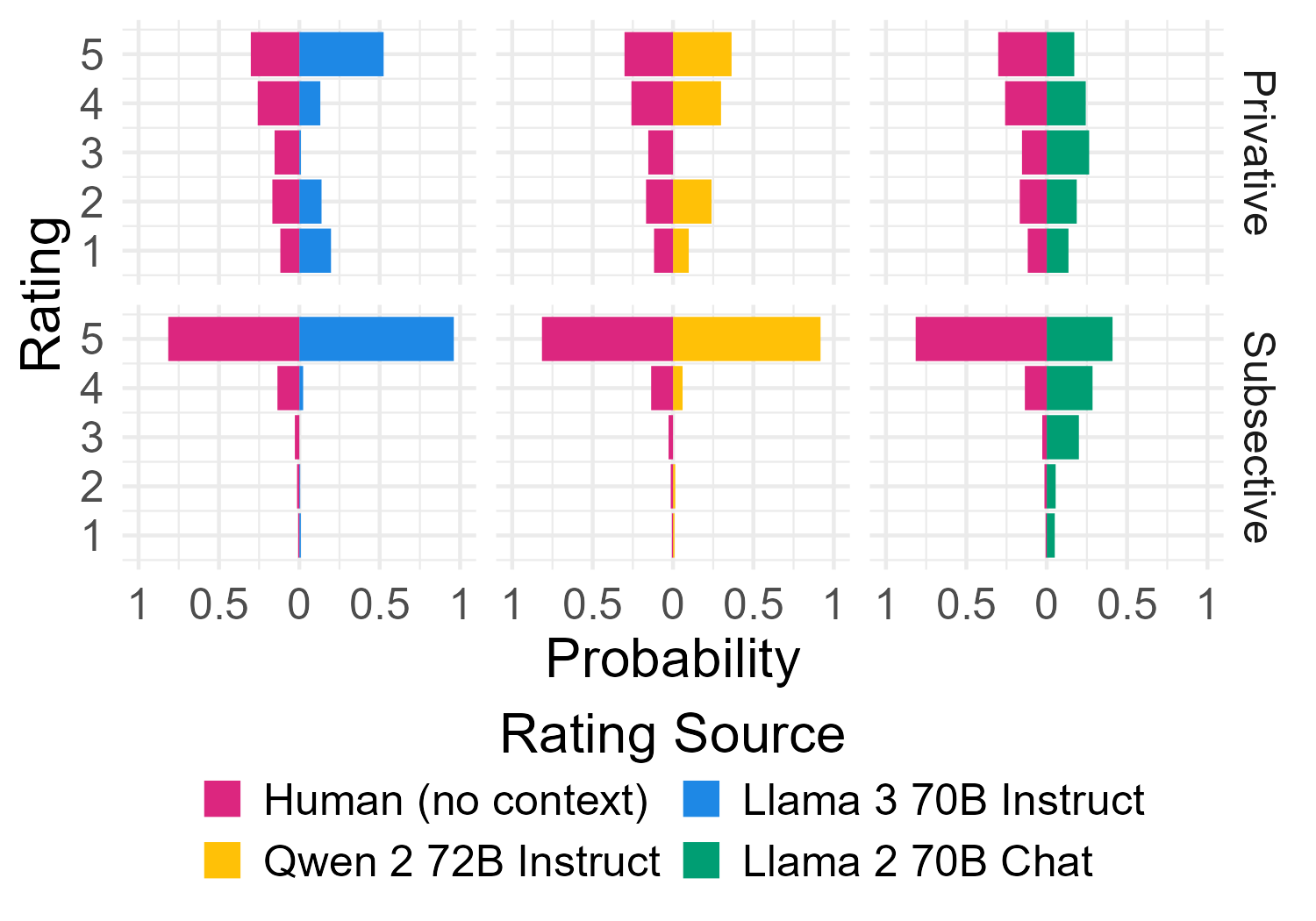}
  \caption{Average log-probability distribution for (typically) subsective vs.~privative adjectives for selected LLMs, compared to the average human distribution.}
  \label{fig:logprob-distributions-by-adjclass}
\end{figure}

\begin{figure*}[t]
  \includegraphics[width=\textwidth]{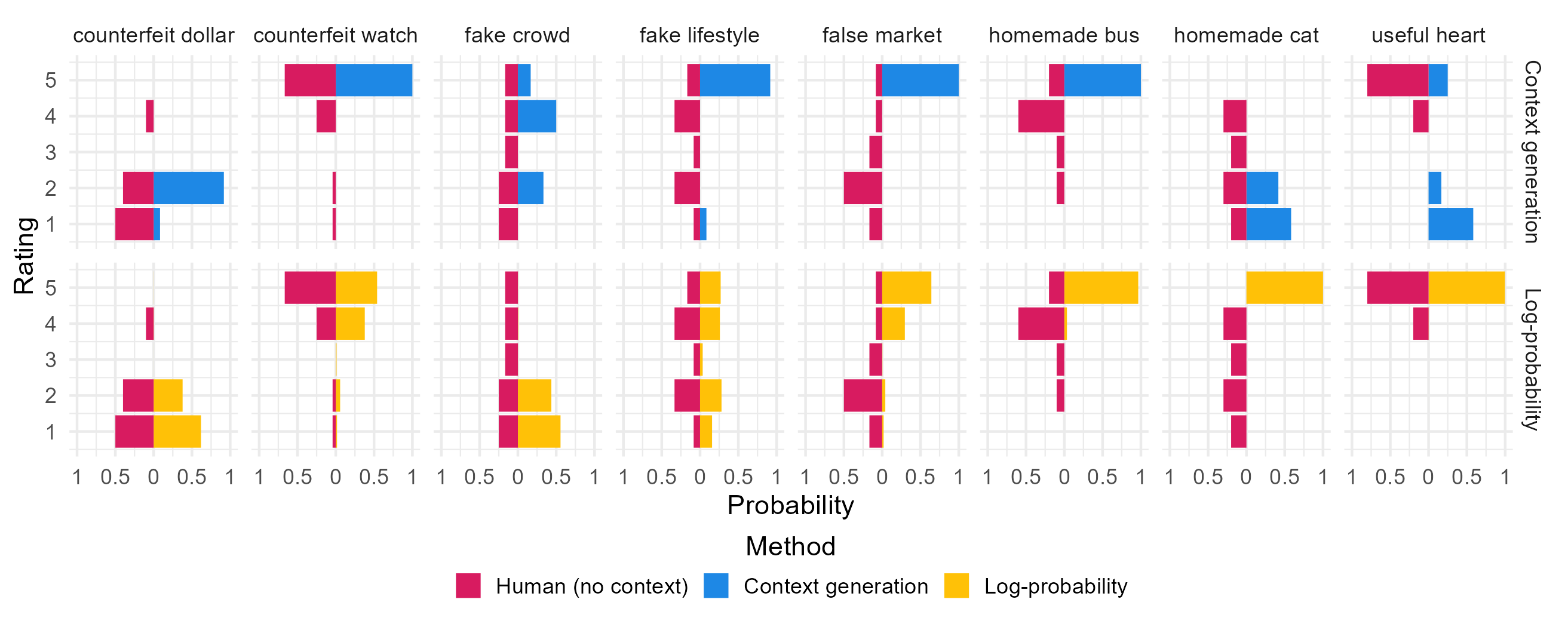}
  \caption{Rating distributions over selected bigrams using log-probabilities vs.~context generation, compared to the human distributions. Both methods successfully capture bigrams like \textit{counterfeit dollar, counterfeit watch}, and both fail for \textit{false market} and \textit{homemade bus}. While the log-probability method fits most bigrams better, such as \textit{fake lifestyle} and \textit{useful heart}, the context generation method is better for \textit{fake crowd} and \textit{homemade cat}.}
  \label{fig:logprob-context-human-bigram-distributions}
\end{figure*}

Table \ref{tab:logprob-js} shows the average Jensen-Shannon divergences,
including a uniform distribution baseline and the ``majority'' baseline reported in Section \ref{sec:acc-1-sd-method}.
Llama 3 70B Instruct shows the lowest average Jensen-Shannon divergence across all bigrams using this method, with an excellent divergence of just 0.08 on (typically) subsective adjectives, matched by Qwen 2 72B. 
However, not all models are able to concentrate enough of their log-probability mass on ``Definitely yes'' for bigrams with subsective adjectives, such as Llama 2 70B Chat, even though
all models
were able to rate these bigrams as subsective when giving a single rating (Table \ref{tab:exp2-1sd}).
Further, all models, especially the larger ones, struggle with subsective-adjective bigrams that humans rate as somewhat privative, such as \textit{homemade cat} or \textit{illegal currency}. 
All models except Qwen 72B Instruct rate \textit{homemade cat} as mostly subsective (5), whereas humans' ratings are distributed evenly from 1-4, and Qwen does not assign enough enough probability mass to these intermediate ratings. 
Figure \ref{fig:logprob-context-human-bigram-distributions} shows the distribution for \textit{homemade cat} for Llama 3 70B Instruct.

For bigrams with privative adjectives, where the inference is much harder to predict, results are still promising, but there is room for improvement. Figure \ref{fig:logprob-distributions-by-adjclass} shows that although Llama 3 70B Instruct has the lowest overall JS divergence for privative adjectives, it rates them as subsective (5, ``Definitely yes'') too often, and neither it nor Qwen 2 72B make human-like use of the ``Unsure'' rating (perhaps a side-effect of their helpfulness training). 
Section \ref{sec:method-comparison} discusses the item-by-item variation in JS divergence in more detail.

More broadly, it is not clear that model log-probabilities  \emph{should} map onto a distribution of how frequent different answers are among humans. 
Method 3 investigates a more sophisticated way of getting a distribution of ratings from LLMs which may map more closely onto the human distribution.

\subsection{Method 3: Context generation} \label{sec:context-generation}

\begin{table}[t]
    \centering
    \begin{tabular}{lccc}
    \toprule
    & \multicolumn{3}{c}{JS Divergence} \\
    Method & Priv. & Subs. & Total \\ \midrule
    Log-probability & 0.26 & 0.08 & 0.17 \\
    Context generation & 0.38 & 0.11 & 0.24 \\
    \bottomrule
    \end{tabular}
    \caption{Jensen-Shannon divergence between per-bigram rating distributions for humans and Llama 3 70B Instruct using the log-probability and context generation methods, for privative vs.~subsective adjectives.}
    \label{tab:exp-3-js-comparison}
\end{table}

\begin{figure}[t]
  \includegraphics[width=\columnwidth]{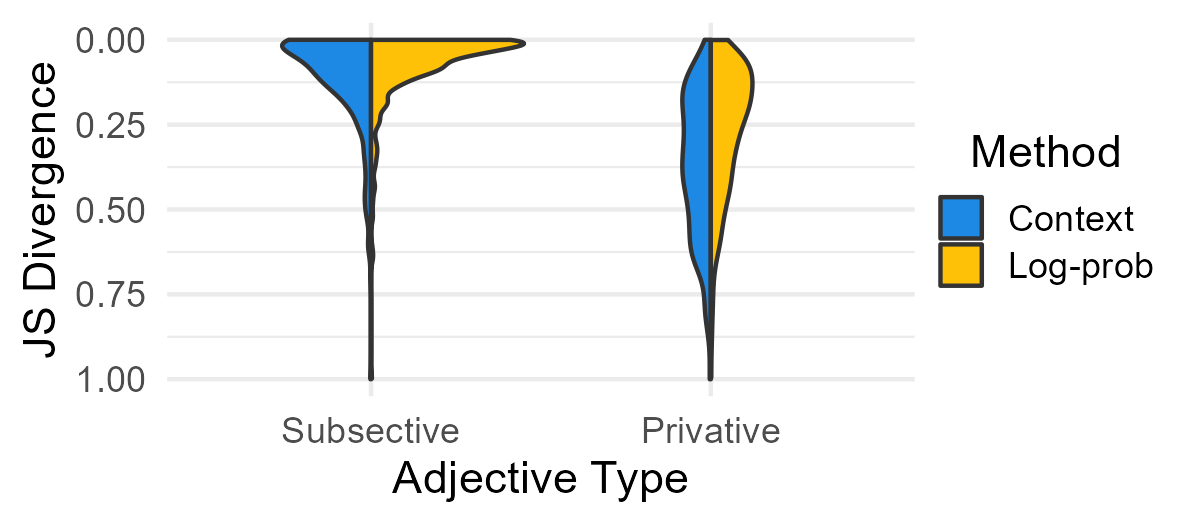}
  \caption{Distribution of per-bigram Jensen-Shannon divergences between the rating distributions for humans and Llama 3 70B when obtained from log-probabilities vs. by generating contexts. 0 indicates perfect overlap, while 1 indicates maximal divergence.}
  \label{fig:js-divergence-split-violin-by-method}
\end{figure}

One source of variation in human ratings is that different humans are likely imagining different instantiations of the bigram, loosely corresponding to the different contexts in Experiment 1 (though likely much less well specified), depending on the priors they each have. They then rate ``Is an AN an N?'' given that imagined context. While some humans may consider multiple instantiations and form a small distribution which informs their single answer, the distribution of ratings primarily arises from consulting a larger population of humans.
A language model, when prompted with no context, instead has a single set of priors and has to estimate the log-probability of each answer given that single set of priors. 
The next method investigates whether we can improve the fit of the model's distribution by having it generate a set of contexts and rate ``Is an AN an N?'' given each context, imitating this aspect of the human variation and thought process.

\subsubsection{Method}

Method 3 asks the model to generate 12 different ``stories'' of 50-100 words involving the target bigram which ``describe the \{bigram\} in detail'', giving three of the contexts used as few-shot prompts for Experiment 1b as examples. 
This is somewhat similar to recent efforts to mimic human survey results by prompting or having the LLM generate personas, then generating data with those personas as context (\citealp{bisbee_synthetic_2024, argyle_out_2023, chan_scaling_2024} i.a.).
Experiment 3a generates all 12 stories in one chat using a temperature of 0.6 (see Appendix \ref{app:generation-prompts} for the prompts). 
Then, Experiment 3b uses the ``In this setting, is an \{adjective\} \{noun\} still a \{noun\}?'' design from Experiment 1b to have the model rate the bigram inference in each of these 12 contexts, yielding 12 ratings per bigram. We can then calculate the Jensen-Shannon divergence between this LLM-generated distribution and the human distribution.

However, 
this method is computationally expensive: it took us ca.~400 GPU-hours with A100s to generate the 12 contexts for our 798 bigrams with Llama 3 70B Instruct. Thus, we conduct this experiment with one model (Llama 3 70B Instruct) and demonstrate this method as a proof of concept.

\subsubsection{Results}

\begin{table*}
\centering
    \begin{tabular}{p{0.47\textwidth}  p{0.47\textwidth}}
    \toprule
         \textbf{Context A} & \textbf{Context B} \\ \midrule
         The new video game, ``Epic Quest,'' was about to be launched, and the developers wanted to create a buzz around it. They decided to stage a fake crowd of fans waiting in line outside the game store on launch night. They \textbf{hired a team of people to dress up in costumes} and hold signs that read ``I've been waiting for 10 hours!'' 
         [\ldots] &
         The small town of Oakdale was hosting its annual Christmas market, but the organizers were worried that not enough people would show up. To create the illusion of a bigger crowd, they \textbf{set up a fake crowd of mannequins} dressed in winter coats and hats, and placed them around the market stalls. [\ldots] \\
         \midrule
         \textbf{Rating}: Definitely yes (subsective) & \textbf{Rating}: Probably not (privative) \\
         \bottomrule
    \end{tabular}
    \caption{Two intuitive stories generated by Llama 3 70B Instruct about a \textit{fake crowd}, which yield privative vs. subsective inferences. We then ask Llama 3 to rate ``Is a fake crowd still a crowd?'' given these contexts.}
    \label{tab:fake-crowd-examples}
\end{table*}

Table \ref{tab:fake-crowd-examples} shows two LLM-written stories illustrating a privative vs. subsective inference for \textit{fake crowd} which successfully capture human intuitions about two kinds of \textit{fake crowd}, showing promise for this method. 
Appendix \ref{app:generated-contexts} contains additional examples of generated contexts.
Overall, however, the Jensen-Shannon divergences in Table \ref{tab:exp-3-js-comparison} show that generating contexts actually fits the human distribution \emph{worse} than just taking the log-probability distribution directly for Llama 3 70B Instruct. 
Using a cut-off of 0.25 for JS divergence, we find that the distributions generated using context generation are reasonably human-like for only 61.4\% of the 798 bigrams, compared to 75.3\% for the log-probability distributions.
One possible explanation is that we are not generating the right kind, or a sufficient diversity, of contexts: we place hardly any constraints on the story generation, but perhaps e.g.~explicitly asking for stories that disambiguate the target inference 
might match human behavior better, since humans see the ``Is an AN an N?'' question when imagining their ``contexts''.
A final point of divergence from humans is Llama 3 70B Instruct's unwillingness to ever use the ``Unsure'' rating, but this may be an issue with this particular model rather than the method itself.
Nevertheless, this method yields well-distributed (often bimodal) rating distributions, as shown in Figure \ref{fig:logprob-context-human-bigram-distributions}, and, while not as close as the log-probability distribution overall, still approximates the human distribution well for many bigrams.

\subsection{Distribution method comparison} \label{sec:method-comparison}

Both methods of generating a distribution are good at capturing the narrow subsective distributions of most subsective adjectives. 
However, both methods struggle for items like \textit{homemade bus}, which humans rate as more privative than subsective despite the typically-subsective adjective, resulting in a thin but long tail of high JS divergences (see Figure \ref{fig:js-divergence-split-violin-by-method}, which compares the distribution of JS divergences per bigram between the two methods). 
Interestingly, the context generation method is better able to capture this partially privative behavior for certain bigrams, such as \textit{homemade cat}, which shows promise: it writes stories focusing around knitted or cardboard cats, matching human intuitions. (This yields a JS divergence of 0.33 compared to 1.00 for the log-prob distribution; see Figure \ref{fig:logprob-context-human-bigram-distributions} and Appendix \ref{app:generated-contexts} for an example context.)

For the typically-privative adjectives which are the primary focus of this paper, the log-probability distributions provide a better fit overall, but not for all bigrams. For 45 of them, such as \textit{fake leg}, context generation provides a better fit: many of the generated stories are about prosthetics, deemed to be \textit{legs} (see Appendix \ref{app:generated-contexts} for an example, and Table \ref{tab:method-comparison-bigram-counts} in the Appendix for more counts).
More broadly, Figure \ref{fig:js-divergence-split-violin-by-method} shows that there is room for improvement for both methods: both have a thick tail of bigrams whose human distributions they do not fit well. 

We fit linear regressions for \texttt{JSDivergence $\sim$ AdjectiveType * HumanMean + HumanSD + BigramFrequency} in R for each method for Llama 3 70B Instruct and find, for each method, a significant negative effect of privative adjective type and human mean on the JS divergence, as well as a significant positive effect of human SD and a significant, positive interaction between adjective type and human mean. We do not find an effect of bigram frequency. 
This is an exciting result, because it shows that Llama 3 70B Instruct is similarly adept at modelling the human distribution of ratings for novel (zero-frequency) bigrams as it is for high-frequency bigrams, suggesting that it can generalize beyond its training data.
Exact coefficients and an effects plot are given in the Appendix in Table \ref{tab:method-regression} and Figure \ref{fig:method-regression-effects}.
Specifically, the negative effect of human mean on typically subsective adjectives supports the qualitative finding that both methods struggle with bigrams involving typically subsective adjectives with low human ratings, i.e.~which have a more privative interpretation. 
Interestingly, we see the same effect for privative adjectives, where JS divergence increases as the human mean decreases (i.e.~as the bigram is rated more privative).
Finally, the significant positive effect of human SD shows that both methods struggle to predict the human distribution as human variation increases.  
These linear regressions achieve an $R^2$ of 0.44 (log-probability) and 0.55 (context generation), meaning there is still variation left unexplained by these factors. 
However, we find no further qualitatively interpretable patterns.

\section{Conclusion}

In this paper, we study whether LLMs can combine adjectives and nouns to yield noun membership inferences both with and without context, for high-frequency and zero-frequency (presumed novel) bigrams. We find that when the inference is determined by context, large, recent LLMs make the expected inferences, while smaller and/or older LLMs only sometimes do so.
All LLMs, even smaller ones, behave similarly for zero-frequency bigrams as they do for high-frequency bigrams, suggesting that they do not rely on having seen them and their associated inference during pretraining. Instead, they evaluate these combinations on-the-fly and can generalize accordingly, as humans do. 

We investigate three methods to evaluate whether LLMs can extend this behavior to a setting where no context is provided, where they either need to fall within the human distribution of ratings or match the whole human distribution. 
While recent LLMs are able to fall within 1 SD of the human mean for up to 95\% of our dataset, this is a very lenient metric.
On our stricter metric of matching the human distribution (either using log-probabilities or via context generation), we find that these LLMs are good at capturing the simple distribution of bigrams like \textit{multicolored flower} and \textit{counterfeit watch}, but are only able to capture up to 75\% of our total dataset. 
LLMs struggle particularly with bigrams with unusual inferences for their adjective, such as \textit{homemade cat}, and with bigrams with high human variance.
Interestingly, however, LLMs are still equally able to capture distributions of novel bigrams in this setting.
In sum, our generalization results are exciting because they show LLMs are generalizing beyond their training data even in this delicate, context-sensitive task, but there is still room for improvement on the task of capturing these inferences overall.
Further, we hope that the methods presented in this paper will inspire future work which moves beyond targeting single correct answers and begins to target human population distributions, from meaning and inferences to wider issues such as opinions and political positions.

\section{Limitations} 

This paper has a number of limitations, being the first exploration in this area. 
Firstly, we only study English adjective-noun inferences. The set of typically-privative adjectives and which inference they trigger with which nouns may vary between languages and cultures. 
Secondly, the dataset for Experiment 1, which determines the inference by providing an appropriate context, is not very large (56 bigram/context pairs), resulting in lower than ideal statistical power. 

Third, for fair comparison to human participants, and since the human data demonstrates that many judgments are non-binary for this task, we use a 5-point Likert scale rather than a binary yes/no question. 
This assumes that the model can use the Likert scale. While there is some previous work which also uses Likert scales with similar models \citep{argyle_out_2023, chuang_beyond_2024, abeysinghe_challenges_2024}, there is no work explicitly showing that models understand Likert scales in general. 
Thus, this design does not let us disentangle whether the model is struggling with the task itself or simply with the use of the Likert scale, despite the 5-shot setting and the use of log-probabilities to enforce use of the scale. This especially applies to smaller models that are known to show weaker instruction-following skills, and to non-instruction-tuned models.
Alternatives such as calculating the surprisal of e.g.~\textit{a fake watch is/is not a watch}, however, have their own issues, since LLMs have been shown to be sometimes insensitive to negation \citep{truong_language_2023,garcia-ferrero_this_2023}.

Fourth, for comparison reasons, we use the same prompts used in the human experiment to evaluate the LLMs, including few-shot examples; however, it is possible that this prompt phrasing and exact choice of few-shot examples introduces artefacts which adversely affect the LLMs' performance and ``masks'' their underlying ability at the task.

Finally, due to time and compute limitations, we were only able to run the context generation method for Llama 3 70B Instruct. In future work, we would like to extend this method to other models and investigate which of the differences we see between the context generation and the log-probability sampling method for Llama 3 70B Instruct should be attributed to the model vs.~to the differences between the methods.

\section*{Acknowledgments}

We would like to thank Naomi Saphra, Jack Merullo, and Lio Wong, as well as all the members of the Harvard Meaning \& Modality lab and the BU tinlab for their helpful feedback. Special thanks to Joshua Martin, whose dissertation on this topic inspired this project. This work was supported by an MBB Graduate Student Research Award from Harvard's Mind, Brain and Behavior Initiative.

\bibliography{anthology,genbench}

\begin{thebibliography}{50}
\providecommand{\natexlab}[1]{#1}

\bibitem[{Abeysinghe and Circi(2024)}]{abeysinghe_challenges_2024}
Bhashithe Abeysinghe and Ruhan Circi. 2024.
\newblock \href {https://doi.org/10.48550/arXiv.2406.03339} {The {Challenges} of {Evaluating} {LLM} {Applications}: {An} {Analysis} of {Automated}, {Human}, and {LLM}-{Based} {Approaches}}.
\newblock \emph{arXiv preprint}.
\newblock ArXiv:2406.03339 [cs].

\bibitem[{Apidianaki and Garí~Soler(2021)}]{apidianaki_all_2021}
Marianna Apidianaki and Aina Garí~Soler. 2021.
\newblock \href {https://doi.org/10.18653/v1/2021.blackboxnlp-1.7} {{ALL} {Dolphins} {Are} {Intelligent} and {SOME} {Are} {Friendly}: {Probing} {BERT} for {Nouns}' {Semantic} {Properties} and their {Prototypicality}}.
\newblock In \emph{Proceedings of the {Fourth} {BlackboxNLP} {Workshop} on {Analyzing} and {Interpreting} {Neural} {Networks} for {NLP}}, pages 79--94, Punta Cana, Dominican Republic. Association for Computational Linguistics.

\bibitem[{Argyle et~al.(2023)Argyle, Busby, Fulda, Gubler, Rytting, and Wingate}]{argyle_out_2023}
Lisa~P. Argyle, Ethan~C. Busby, Nancy Fulda, Joshua~R. Gubler, Christopher Rytting, and David Wingate. 2023.
\newblock \href {https://doi.org/10.1017/pan.2023.2} {Out of {One}, {Many}: {Using} {Language} {Models} to {Simulate} {Human} {Samples}}.
\newblock \emph{Political Analysis}, 31(3):337--351.

\bibitem[{Bertolini et~al.(2022)Bertolini, Weeds, and Weir}]{bertolini_testing_2022}
Lorenzo Bertolini, Julie Weeds, and David Weir. 2022.
\newblock \href {https://aclanthology.org/2022.coling-1.359} {Testing {Large} {Language} {Models} on {Compositionality} and {Inference} with {Phrase}-{Level} {Adjective}-{Noun} {Entailment}}.
\newblock In \emph{Proceedings of the 29th {International} {Conference} on {Computational} {Linguistics}}, pages 4084--4100, Gyeongju, Republic of Korea. International Committee on Computational Linguistics.

\bibitem[{Bisbee et~al.(2024)Bisbee, Clinton, Dorff, Kenkel, and Larson}]{bisbee_synthetic_2024}
James Bisbee, Joshua~D. Clinton, Cassy Dorff, Brenton Kenkel, and Jennifer~M. Larson. 2024.
\newblock \href {https://doi.org/10.1017/pan.2024.5} {Synthetic {Replacements} for {Human} {Survey} {Data}? {The} {Perils} of {Large} {Language} {Models}}.
\newblock \emph{Political Analysis}, pages 1--16.

\bibitem[{Boleda et~al.(2013)Boleda, Baroni, Pham, and McNally}]{boleda_intensionality_2013}
Gemma Boleda, Marco Baroni, The~Nghia Pham, and Louise McNally. 2013.
\newblock \href {https://aclanthology.org/W13-0104} {Intensionality was only alleged: {On} adjective-noun composition in distributional semantics}.
\newblock In \emph{Proceedings of the 10th {International} {Conference} on {Computational} {Semantics} ({IWCS} 2013) – {Long} {Papers}}, pages 35--46, Potsdam, Germany. Association for Computational Linguistics.

\bibitem[{Boleda et~al.(2012)Boleda, Vecchi, Cornudella, and McNally}]{boleda_first_2012}
Gemma Boleda, Eva~Maria Vecchi, Miquel Cornudella, and Louise McNally. 2012.
\newblock \href {https://aclanthology.org/D12-1112} {First {Order} vs. {Higher} {Order} {Modification} in {Distributional} {Semantics}}.
\newblock In \emph{Proceedings of the 2012 {Joint} {Conference} on {Empirical} {Methods} in {Natural} {Language} {Processing} and {Computational} {Natural} {Language} {Learning}}, pages 1223--1233, Jeju Island, Korea. Association for Computational Linguistics.

\bibitem[{Brown et~al.(2020)Brown, Mann, Ryder, Subbiah, Kaplan, Dhariwal, Neelakantan, Shyam, Sastry, Askell, Agarwal, Herbert-Voss, Krueger, Henighan, Child, Ramesh, Ziegler, Wu, Winter, Hesse, Chen, Sigler, Litwin, Gray, Chess, Clark, Berner, McCandlish, Radford, Sutskever, and Amodei}]{brown_language_2020}
Tom~B. Brown, Benjamin Mann, Nick Ryder, Melanie Subbiah, Jared Kaplan, Prafulla Dhariwal, Arvind Neelakantan, Pranav Shyam, Girish Sastry, Amanda Askell, Sandhini Agarwal, Ariel Herbert-Voss, Gretchen Krueger, Tom Henighan, Rewon Child, Aditya Ramesh, Daniel~M. Ziegler, Jeffrey Wu, Clemens Winter, Christopher Hesse, Mark Chen, Eric Sigler, Mateusz Litwin, Scott Gray, Benjamin Chess, Jack Clark, Christopher Berner, Sam McCandlish, Alec Radford, Ilya Sutskever, and Dario Amodei. 2020.
\newblock \href {http://arxiv.org/abs/2005.14165} {Language {Models} are {Few}-{Shot} {Learners}}.
\newblock \emph{arXiv:2005.14165 [cs]}.
\newblock ArXiv: 2005.14165.

\bibitem[{Cappelle et~al.(2018)Cappelle, Pascal, and Keller}]{cappelle_facing_2018}
Bert Cappelle, Denis Pascal, and Mikaela Keller. 2018.
\newblock \href {https://doi.org/10.1515/gcla-2018-0002} {Facing the facts of fake: {A} distributional semantics and corpus annotation approach}.
\newblock \emph{Yearbook of the German Cognitive Linguistics Association}, 6(1):9--42.

\bibitem[{Chan et~al.(2024)Chan, Wang, Yu, Mi, and Yu}]{chan_scaling_2024}
Xin Chan, Xiaoyang Wang, Dian Yu, Haitao Mi, and Dong Yu. 2024.
\newblock \href {http://arxiv.org/abs/2406.20094} {Scaling {Synthetic} {Data} {Creation} with 1,000,000,000 {Personas}}.
\newblock \emph{arXiv preprint}.
\newblock ArXiv:2406.20094 [cs].

\bibitem[{Christensen(2022)}]{christensen_ordinalregression_2022}
R.~H.~B. Christensen. 2022.
\newblock ordinal—{Regression} {Models} for {Ordinal} {Data}.
\newblock R package version 2022.11-16. https://CRAN.R-project.org/package=ordinal.

\bibitem[{Chuang et~al.(2024)Chuang, Studdiford, Nirunwiroj, Goyal, Frigo, Yang, Shah, Hu, and Rogers}]{chuang_beyond_2024}
Yun-Shiuan Chuang, Zach Studdiford, Krirk Nirunwiroj, Agam Goyal, Vincent~V. Frigo, Sijia Yang, Dhavan Shah, Junjie Hu, and Timothy~T. Rogers. 2024.
\newblock \href {https://doi.org/10.48550/arXiv.2406.17232} {Beyond {Demographics}: {Aligning} {Role}-playing {LLM}-based {Agents} {Using} {Human} {Belief} {Networks}}.
\newblock \emph{arXiv preprint}.
\newblock ArXiv:2406.17232 [cs].

\bibitem[{Coil and Shwartz(2023)}]{coil_chocolate_2023}
Albert Coil and Vered Shwartz. 2023.
\newblock \href {https://doi.org/10.18653/v1/2023.findings-acl.169} {From chocolate bunny to chocolate crocodile: {Do} {Language} {Models} {Understand} {Noun} {Compounds}?}
\newblock In \emph{Findings of the {Association} for {Computational} {Linguistics}: {ACL} 2023}, pages 2698--2710, Toronto, Canada. Association for Computational Linguistics.

\bibitem[{Devlin et~al.(2019)Devlin, Chang, Lee, and Toutanova}]{devlin_bert_2019}
Jacob Devlin, Ming-Wei Chang, Kenton Lee, and Kristina Toutanova. 2019.
\newblock \href {https://doi.org/10.18653/v1/N19-1423} {{BERT}: {Pre}-training of {Deep} {Bidirectional} {Transformers} for {Language} {Understanding}}.
\newblock In \emph{Proceedings of the 2019 {Conference} of the {North} {American} {Chapter} of the {Association} for {Computational} {Linguistics}: {Human} {Language} {Technologies}, {Volume} 1 ({Long} and {Short} {Papers})}, pages 4171--4186, Minneapolis, Minnesota. Association for Computational Linguistics.

\bibitem[{Do and Pavlick(2021)}]{do_are_2021}
Nam Do and Ellie Pavlick. 2021.
\newblock \href {https://doi.org/10.18653/v1/2021.findings-acl.181} {Are {Rotten} {Apples} {Edible}? {Challenging} {Commonsense} {Inference} {Ability} with {Exceptions}}.
\newblock In \emph{Findings of the {Association} for {Computational} {Linguistics}: {ACL}-{IJCNLP} 2021}, pages 2061--2073, Online. Association for Computational Linguistics.

\bibitem[{Dubey et~al.(2024)Dubey, Jauhri, Pandey, Kadian, Al-Dahle, Letman, Mathur, Schelten, Yang, Fan, and {et al.}}]{dubey_llama_2024}
Abhimanyu Dubey, Abhinav Jauhri, Abhinav Pandey, Abhishek Kadian, Ahmad Al-Dahle, Aiesha Letman, Akhil Mathur, Alan Schelten, Amy Yang, Angela Fan, and {}~{et al.} 2024.
\newblock \href {https://doi.org/10.48550/arXiv.2407.21783} {The {Llama} 3 {Herd} of {Models}}.
\newblock \emph{arXiv preprint}.
\newblock ArXiv:2407.21783 [cs].

\bibitem[{Ettinger(2020)}]{ettinger_what_2020}
Allyson Ettinger. 2020.
\newblock \href {https://doi.org/10.1162/tacl_a_00298} {What {BERT} {Is} {Not}: {Lessons} from a {New} {Suite} of {Psycholinguistic} {Diagnostics} for {Language} {Models}}.
\newblock \emph{Transactions of the Association for Computational Linguistics}, 8:34--48.

\bibitem[{García-Ferrero et~al.(2023)García-Ferrero, Altuna, Álvez, Gonzalez-Dios, and Rigau}]{garcia-ferrero_this_2023}
Iker García-Ferrero, Begoña Altuna, Javier Álvez, Itziar Gonzalez-Dios, and German Rigau. 2023.
\newblock \href {https://doi.org/10.48550/arXiv.2310.15941} {This is not a {Dataset}: {A} {Large} {Negation} {Benchmark} to {Challenge} {Large} {Language} {Models}}.
\newblock \emph{arXiv preprint}.
\newblock ArXiv:2310.15941 [cs].

\bibitem[{Goodale and Mascarenhas(2023)}]{goodale_systematic_2023}
Michael Goodale and Salvador Mascarenhas. 2023.
\newblock \href {https://lingbuzz.net/lingbuzz/007644} {Systematic polysemy in adjective-noun combination in contextual word embeddings}.
\newblock LingBuzz Published In:.

\bibitem[{Guevara(2010)}]{guevara_regression_2010}
Emiliano Guevara. 2010.
\newblock \href {https://aclanthology.org/W10-2805} {A {Regression} {Model} of {Adjective}-{Noun} {Compositionality} in {Distributional} {Semantics}}.
\newblock In \emph{Proceedings of the 2010 {Workshop} on {GEometrical} {Models} of {Natural} {Language} {Semantics}}, pages 33--37, Uppsala, Sweden. Association for Computational Linguistics.

\bibitem[{Hacken(2016)}]{hacken_semantics_2016}
Pius~ten Hacken. 2016.
\newblock \emph{The {Semantics} of {Compounding}}.
\newblock Cambridge University Press.
\newblock Google-Books-ID: esLgCwAAQBAJ.

\bibitem[{Hanna and Mareček(2021)}]{hanna_analyzing_2021}
Michael Hanna and David Mareček. 2021.
\newblock \href {https://doi.org/10.18653/v1/2021.blackboxnlp-1.20} {Analyzing {BERT}'s {Knowledge} of {Hypernymy} via {Prompting}}.
\newblock In \emph{Proceedings of the {Fourth} {BlackboxNLP} {Workshop} on {Analyzing} and {Interpreting} {Neural} {Networks} for {NLP}}, pages 275--282, Punta Cana, Dominican Republic. Association for Computational Linguistics.

\bibitem[{Hartung et~al.(2017)Hartung, Kaupmann, Jebbara, and Cimiano}]{hartung_learning_2017}
Matthias Hartung, Fabian Kaupmann, Soufian Jebbara, and Philipp Cimiano. 2017.
\newblock \href {https://aclanthology.org/E17-1006} {Learning {Compositionality} {Functions} on {Word} {Embeddings} for {Modelling} {Attribute} {Meaning} in {Adjective}-{Noun} {Phrases}}.
\newblock In \emph{Proceedings of the 15th {Conference} of the {European} {Chapter} of the {Association} for {Computational} {Linguistics}: {Volume} 1, {Long} {Papers}}, pages 54--64, Valencia, Spain. Association for Computational Linguistics.

\bibitem[{Jiang et~al.(2024)Jiang, Sablayrolles, Roux, Mensch, Savary, Bamford, Chaplot, Casas, Hanna, Bressand, Lengyel, Bour, Lample, Lavaud, Saulnier, Lachaux, Stock, Subramanian, Yang, Antoniak, Scao, Gervet, Lavril, Wang, Lacroix, and Sayed}]{jiang_mixtral_2024}
Albert~Q. Jiang, Alexandre Sablayrolles, Antoine Roux, Arthur Mensch, Blanche Savary, Chris Bamford, Devendra~Singh Chaplot, Diego de~las Casas, Emma~Bou Hanna, Florian Bressand, Gianna Lengyel, Guillaume Bour, Guillaume Lample, Lélio~Renard Lavaud, Lucile Saulnier, Marie-Anne Lachaux, Pierre Stock, Sandeep Subramanian, Sophia Yang, Szymon Antoniak, Teven~Le Scao, Théophile Gervet, Thibaut Lavril, Thomas Wang, Timothée Lacroix, and William~El Sayed. 2024.
\newblock \href {https://doi.org/10.48550/arXiv.2401.04088} {Mixtral of {Experts}}.
\newblock \emph{arXiv preprint}.
\newblock ArXiv:2401.04088 [cs].

\bibitem[{Kim et~al.(2022)Kim, Linzen, and Smolensky}]{kim_uncontrolled_2022}
Najoung Kim, Tal Linzen, and Paul Smolensky. 2022.
\newblock \href {https://doi.org/10.48550/arXiv.2212.10769} {Uncontrolled {Lexical} {Exposure} {Leads} to {Overestimation} of {Compositional} {Generalization} in {Pretrained} {Models}}.
\newblock \emph{arXiv preprint}.
\newblock ArXiv:2212.10769 [cs].

\bibitem[{Lalisse and Asudeh(2015)}]{lalisse_distinguishing_2015}
Matthias Lalisse and Ash Asudeh. 2015.
\newblock \href {https://pages.jh.edu/mlaliss1/media/MLalisse_thesis.pdf} {Distinguishing intersective and non-intersective adjectives in compositional distributional semantics}.
\newblock Master's thesis, University of Oxford, Oxford.

\bibitem[{Liu et~al.(2019)Liu, Ott, Goyal, Du, Joshi, Chen, Levy, Lewis, Zettlemoyer, and Stoyanov}]{liu_roberta_2019}
Yinhan Liu, Myle Ott, Naman Goyal, Jingfei Du, Mandar Joshi, Danqi Chen, Omer Levy, Mike Lewis, Luke Zettlemoyer, and Veselin Stoyanov. 2019.
\newblock \href {http://arxiv.org/abs/1907.11692} {{RoBERTa}: {A} {Robustly} {Optimized} {BERT} {Pretraining} {Approach}}.
\newblock \emph{arXiv:1907.11692 [cs]}.
\newblock ArXiv: 1907.11692.

\bibitem[{Lyu et~al.(2022)Lyu, Zheng, Li, Zhang, Apidianaki, and Callison-Burch}]{lyu_is_2022}
Qing Lyu, Hua Zheng, Daoxin Li, Li~Zhang, Marianna Apidianaki, and Chris Callison-Burch. 2022.
\newblock \href {https://doi.org/10.48550/arXiv.2112.08326} {Is "{My} {Favorite} {New} {Movie}" {My} {Favorite} {Movie}? {Probing} the {Understanding} of {Recursive} {Noun} {Phrases}}.
\newblock \emph{arXiv preprint}.

\bibitem[{Martin(2022)}]{martin_compositional_2022}
Joshua Martin. 2022.
\newblock \href {https://www.proquest.com/docview/2681380157/abstract/58A63B8C3E6548AEPQ/1} {\emph{Compositional {Routes} to ({Non}){Intersectivity}}}.
\newblock Ph.{D}., Harvard University, United States -- Massachusetts.

\bibitem[{{MosaicML NLP Team}(2023)}]{MosaicML2023Introducing}
{MosaicML NLP Team}. 2023.
\newblock \href {www.mosaicml.com/blog/mpt-30b} {Introducing mpt-30b: Raising the bar for open-source foundation models}.

\bibitem[{Moskvoretskii et~al.(2024)Moskvoretskii, Panchenko, and Nikishina}]{moskvoretskii_are_2024}
Viktor Moskvoretskii, Alexander Panchenko, and Irina Nikishina. 2024.
\newblock \href {https://aclanthology.org/2024.lrec-main.133} {Are {Large} {Language} {Models} {Good} at {Lexical} {Semantics}? {A} {Case} of {Taxonomy} {Learning}}.
\newblock In \emph{Proceedings of the 2024 {Joint} {International} {Conference} on {Computational} {Linguistics}, {Language} {Resources} and {Evaluation} ({LREC}-{COLING} 2024)}, pages 1498--1510, Torino, Italia. ELRA and ICCL.

\bibitem[{Nayak et~al.(2014)Nayak, Kowarsky, Angeli, and Manning}]{nayak_dictionary_2014}
Neha Nayak, Mark Kowarsky, Gabor Angeli, and Christopher~D. Manning. 2014.
\newblock \href {https://www-cs.stanford.edu/~angeli/papers/2014-tr-adjectives.pdf} {A {Dictionary} of {Nonsubsective} {Adjectives}}.
\newblock Technical Report CSTR 2014-04, Department of Computer Science, Stanford University.

\bibitem[{Nikishina et~al.(2023)Nikishina, Chernomorchenko, Demidova, Panchenko, and Biemann}]{nikishina_predicting_2023}
Irina Nikishina, Polina Chernomorchenko, Anastasiia Demidova, Alexander Panchenko, and Chris Biemann. 2023.
\newblock \href {https://doi.org/10.18653/v1/2023.findings-ijcnlp.12} {Predicting {Terms} in {IS}-{A} {Relations} with {Pre}-trained {Transformers}}.
\newblock In \emph{Findings of the {Association} for {Computational} {Linguistics}: {IJCNLP}-{AACL} 2023 ({Findings})}, pages 134--148, Nusa Dua, Bali. Association for Computational Linguistics.

\bibitem[{OpenAI(2024)}]{openai_gpt-4_2024}
OpenAI. 2024.
\newblock \href {https://doi.org/10.48550/arXiv.2303.08774} {{GPT}-4 {Technical} {Report}}.
\newblock \emph{arXiv preprint}.
\newblock ArXiv:2303.08774 [cs].

\bibitem[{Ormerod et~al.(2024)Ormerod, del Rincón, and Devereux}]{ormerod_how_2024}
Mark Ormerod, Jesús~Martínez del Rincón, and Barry Devereux. 2024.
\newblock \href {https://doi.org/10.1162/coli_a_00495} {How {Is} a “{Kitchen} {Chair}” like a “{Farm} {Horse}”? {Exploring} the {Representation} of {Noun}-{Noun} {Compound} {Semantics} in {Transformer}-based {Language} {Models}}.
\newblock \emph{Computational Linguistics}, 50(1):49--81.
\newblock Place: Cambridge, MA Publisher: MIT Press.

\bibitem[{Pavlick and Callison-Burch(2016{\natexlab{a}})}]{pavlick_most_2016}
Ellie Pavlick and Chris Callison-Burch. 2016{\natexlab{a}}.
\newblock \href {https://doi.org/10.18653/v1/P16-1204} {Most “babies” are “little” and most “problems” are “huge”: {Compositional} {Entailment} in {Adjective}-{Nouns}}.
\newblock In \emph{Proceedings of the 54th {Annual} {Meeting} of the {Association} for {Computational} {Linguistics} ({Volume} 1: {Long} {Papers})}, pages 2164--2173, Berlin, Germany. Association for Computational Linguistics.

\bibitem[{Pavlick and Callison-Burch(2016{\natexlab{b}})}]{pavlick_so-called_2016}
Ellie Pavlick and Chris Callison-Burch. 2016{\natexlab{b}}.
\newblock \href {https://doi.org/10.18653/v1/S16-2014} {So-{Called} {Non}-{Subsective} {Adjectives}}.
\newblock In \emph{Proceedings of the {Fifth} {Joint} {Conference} on {Lexical} and {Computational} {Semantics}}, pages 114--119, Berlin, Germany. Association for Computational Linguistics.

\bibitem[{Pennington et~al.(2014)Pennington, Socher, and Manning}]{pennington_glove_2014}
Jeffrey Pennington, Richard Socher, and Christopher Manning. 2014.
\newblock \href {https://doi.org/10.3115/v1/D14-1162} {{GloVe}: {Global} {Vectors} for {Word} {Representation}}.
\newblock In \emph{Proceedings of the 2014 {Conference} on {Empirical} {Methods} in {Natural} {Language} {Processing} ({EMNLP})}, pages 1532--1543, Doha, Qatar. Association for Computational Linguistics.

\bibitem[{{R Core Team}(2023)}]{r_core_team_r_2023}
{R Core Team}. 2023.
\newblock \href {https://www.R-project.org/} {R: {A} {Language} and {Environment} for {Statistical} {Computing}}.

\bibitem[{Radford et~al.(2019)Radford, Wu, Child, Luan, Amodei, and Sutskever}]{radford_language_2019}
Alec Radford, Jeffrey Wu, Rewon Child, David Luan, Dario Amodei, and Ilya Sutskever. 2019.
\newblock \href {https://cdn.openai.com/better-language-models/language_models_are_unsupervised_multitask_learners.pdf} {Language {Models} are {Unsupervised} {Multitask} {Learners}}.
\newblock page~24.

\bibitem[{Raffel et~al.(2020)Raffel, Shazeer, Roberts, Lee, Narang, Matena, Zhou, Li, and Liu}]{raffel_exploring_2020}
Colin Raffel, Noam Shazeer, Adam Roberts, Katherine Lee, Sharan Narang, Michael Matena, Yanqi Zhou, Wei Li, and Peter~J. Liu. 2020.
\newblock \href {http://jmlr.org/papers/v21/20-074.html} {Exploring the {Limits} of {Transfer} {Learning} with a {Unified} {Text}-to-{Text} {Transformer}}.
\newblock \emph{Journal of Machine Learning Research}, 21(140):1--67.

\bibitem[{Rambelli et~al.(2024)Rambelli, Chersoni, Collacciani, and Bolognesi}]{rambelli_can_2024}
Giulia Rambelli, Emmanuele Chersoni, Claudia Collacciani, and Marianna Bolognesi. 2024.
\newblock \href {https://analogy-angle.github.io/assets/Rambelli_2.pdf} {Can {Large} {Language} {Models} {Interpret} {Noun}-{Noun} {Compounds}? {A} {Linguistically}-{Motivated} {Study} on {Lexicalized} and {Novel} {Compounds}}.
\newblock In \emph{The {First} {Workshop} on {Analogical} {Abstraction} in {Cognition}, {Perception}, and {Language} ({Analogy}-{ANGLE})}.

\bibitem[{Ravichander et~al.(2020)Ravichander, Hovy, Suleman, Trischler, and Cheung}]{ravichander_systematicity_2020}
Abhilasha Ravichander, Eduard Hovy, Kaheer Suleman, Adam Trischler, and Jackie Chi~Kit Cheung. 2020.
\newblock \href {https://aclanthology.org/2020.starsem-1.10} {On the {Systematicity} of {Probing} {Contextualized} {Word} {Representations}: {The} {Case} of {Hypernymy} in {BERT}}.
\newblock In \emph{Proceedings of the {Ninth} {Joint} {Conference} on {Lexical} and {Computational} {Semantics}}, pages 88--102, Barcelona, Spain (Online). Association for Computational Linguistics.

\bibitem[{Ross et~al.(2024)Ross, Kim, and Davidson}]{ross_fake_2024}
Hayley Ross, Najoung Kim, and Kathryn Davidson. 2024.
\newblock \href {https://lingbuzz.net/lingbuzz/008012} {Fake reefs are sometimes reefs and sometimes not, but are always compositional}.
\newblock \emph{to appear in Experiments in Linguistic Meaning}, 3.

\bibitem[{Sathe et~al.(2024)Sathe, Fedorenko, and Zaslavsky}]{sathe_language_2024}
Aalok Sathe, Evelina Fedorenko, and Noga Zaslavsky. 2024.
\newblock \href {https://escholarship.org/uc/item/0qd3662b} {Language use is only sparsely compositional: {The} case of {English} adjective-noun phrases in humans and large language models}.
\newblock \emph{Proceedings of the Annual Meeting of the Cognitive Science Society}, 46(0).

\bibitem[{Touvron et~al.(2023)Touvron, Martin, Stone, and {et al.}}]{touvron_llama_2023-1}
Hugo Touvron, Louis Martin, Kevin Stone, and {}~{et al.} 2023.
\newblock \href {https://doi.org/10.48550/arXiv.2307.09288} {Llama 2: {Open} {Foundation} and {Fine}-{Tuned} {Chat} {Models}}.
\newblock \emph{arXiv preprint}.

\bibitem[{Truong et~al.()Truong, Baldwin, Verspoor, and Cohn}]{truong_language_2023}
Thinh~Hung Truong, Timothy Baldwin, Karin Verspoor, and Trevor Cohn.
\newblock \href {https://doi.org/10.48550/arXiv.2306.08189} {Language models are not naysayers: An analysis of language models on negation benchmarks}.
\newblock \emph{Preprint}, arxiv:2306.08189 [cs].

\bibitem[{Vecchi et~al.(2017)Vecchi, Marelli, Zamparelli, and Baroni}]{vecchi_spicy_2017}
Eva~M. Vecchi, Marco Marelli, Roberto Zamparelli, and Marco Baroni. 2017.
\newblock \href {https://doi.org/10.1111/cogs.12330} {Spicy {Adjectives} and {Nominal} {Donkeys}: {Capturing} {Semantic} {Deviance} {Using} {Compositionality} in {Distributional} {Spaces}}.
\newblock \emph{Cognitive Science}, 41(1):102--136.

\bibitem[{Yang et~al.(2024)Yang, Yang, Hui, Zheng, Yu, Zhou, Li, Li, Liu, Huang, and {et al.}}]{yang_qwen2_2024}
An~Yang, Baosong Yang, Binyuan Hui, Bo~Zheng, Bowen Yu, Chang Zhou, Chengpeng Li, Chengyuan Li, Dayiheng Liu, Fei Huang, and {}~{et al.} 2024.
\newblock \href {https://doi.org/10.48550/arXiv.2407.10671} {Qwen2 {Technical} {Report}}.
\newblock \emph{arXiv preprint}.
\newblock ArXiv:2407.10671 [cs].

\bibitem[{Yu and Ettinger(2020)}]{yu_assessing_2020}
Lang Yu and Allyson Ettinger. 2020.
\newblock \href {https://doi.org/10.18653/v1/2020.emnlp-main.397} {Assessing {Phrasal} {Representation} and {Composition} in {Transformers}}.
\newblock In \emph{Proceedings of the 2020 {Conference} on {Empirical} {Methods} in {Natural} {Language} {Processing} ({EMNLP})}, pages 4896--4907, Online. Association for Computational Linguistics.

\end{thebibliography}

\appendix

\section{Experiment 1a: No-context baseline} \label{app:exp1a}

\begin{figure}[t]
  \includegraphics[width=\columnwidth]{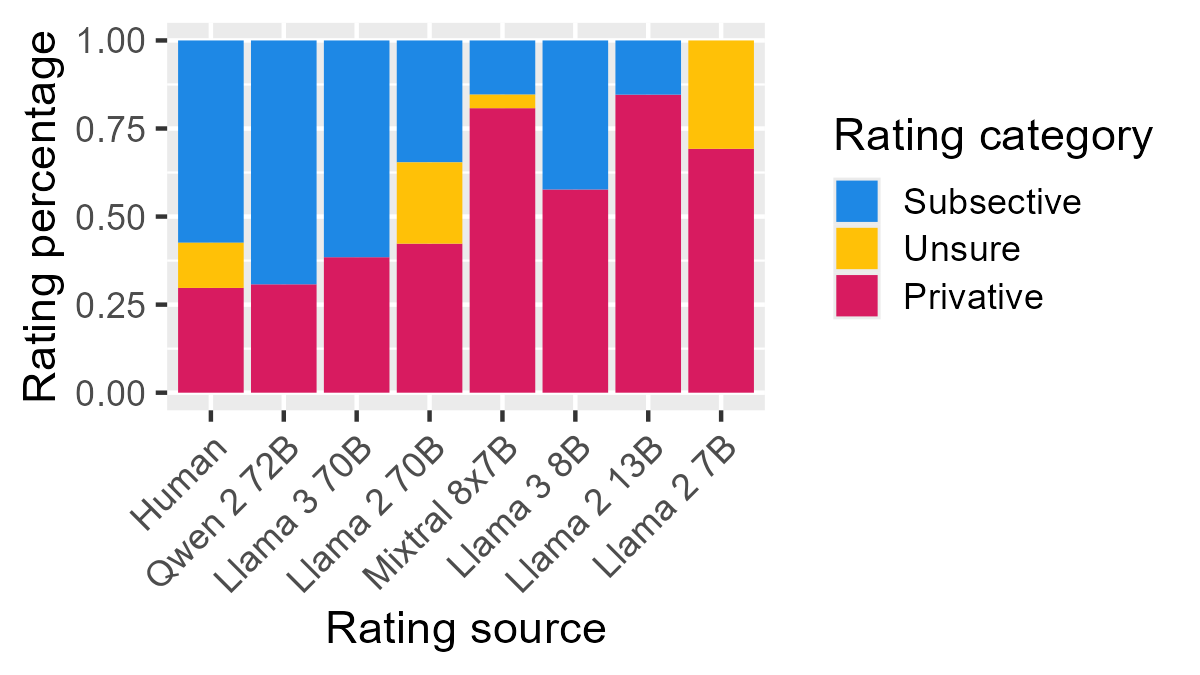}
  \caption{Percentage of privative vs.~subsective inferences for bigrams in Experiment 1a (no context) for each model (``Instruct/Chat'' omitted for brevity). 
  }
  \label{fig:exp1a-no-context}
\end{figure}

Figure \ref{fig:exp1a-no-context} shows the proportions of bigram ratings which are privative (``Definitely not'' or ``Probably not''), subsective (``Definitely yes'' or ``Probably yes'') or neither (``Unsure'') when the LLM (or human; 12 ratings/bigram for humans) rates these 28 bigrams without context in Experiment 1a. The exact proportions are shown in Table \ref{tab:exp1a-no-context}.
Each LLM brings its own bias: Llama 3 70B is evenly balanced, while Qwen 2 72B favors subsective ratings and Mixtral 8x7B favors privative ratings.
For humans, all of these bigrams have high variance when rated in this out-of-context setting, 
which these percentages do not reflect. 

\begin{table}[t]
    \centering
    \begin{tabular}{lccc}
    \toprule
    Model & priv. & subs. & unsure \\ \midrule
    Human & 29.7\% & 57.4\% & 12.8\% \\
    Qwen 2 72B Instruct & 30.8\% & 69.2\%  & 0.0\%  \\
    Llama 3 70B Instruct & 46.2\% & 53.8\% & 0.0\% \\
    Llama 2 70B Chat & 42.3\% & 34.6\% & 23.1\% \\
    Mixtral 7x8B Instruct & 80.8\% & 15.4\% & 3.9\%  \\
    Llama 3 8B Instruct & 57.7\% & 42.3\% & 0.0\% \\
    Llama 2 13B Chat & 84.6\% & 15.4\% & 0.0\% \\
    Llama 2 7B Chat & 69.2\% & 0.0\% & 30.8\% \\ 
    Llama 3 70B & 69.2\% & 30.8\%  &  0.0\%  \\
    Llama 3 8B & 38.5\% & 53.8\%  & 7.7\%   \\
    Llama 2 70B & 19.2\% & 19.2\%  & 61.5\% \\
    Llama 2 13B & 0.0\% &  0.0\% & 100\% \\
    Llama 2 7B & 88.5\% & 11.5\%  & 0.0\%  \\ 
    \bottomrule
    \end{tabular}
    \caption{Percentage of privative vs.~subsective inferences for bigrams in Experiment 1a (no context, 5-shot).}
    \label{tab:exp1a-no-context}
\end{table}

\section{Data and prompts} 

\subsection{Context examples from Ross et al. (2024)} \label{app:context-examples}

To illustrate that the same bigram may be privative or subsective in different contexts, we provide the two contexts for \textit{fake concert} written by \citet{ross_fake_2024} in Table \ref{tab:fake-concert-examples}.

\begin{table*}
\centering
    \begin{tabular}{p{0.47\textwidth}  p{0.47\textwidth}}
    \toprule
         \textbf{Privative-biased context:} & \textbf{Subsective-biased context:} \\ \midrule
         A well-known band gets into trouble when it emerges that they included a \textbf{fake concert} in their tax returns, which they claim had huge financial losses (letting them get away with paying very low taxes), but which never actually happened. &
         A political party disguises a fundraiser as a concert so that they can hold it at a venue where political rallies aren't allowed. They even hire an up-and-coming band to sing at the event. The \textbf{fake concert} is a great success and the attendees enjoy the music as well as networking with the political candidates.\\
         \bottomrule
    \end{tabular}
    \caption{The two contexts in \citet{ross_fake_2024} for \textit{fake concert}, which bias humans towards a privative vs.~subsective rating respectively. Contexts are carefully constructed to determine the inference without explicitly stating it or implying it through value judgments.}
    \label{tab:fake-concert-examples}
\end{table*}

\subsection {Few-shot examples} \label{app:few-shot-examples}

The few-shot prompts for Experiment 1b and Experiment 3b---5 bigrams with contexts, one for each rating---are shown in Table \ref{tab:fewshot-contexts}.
For the chat/instruction-tuned models, we format the few-shot prompt as a conversation between the assistant and the user, where each context and question is provided by the user and the assistant provides each answer (without the ``Answer'' prefix). For the base models, we concatenate the few-shot examples and use the ``Answer'' prefix to indicate the answer. Humans see only the first three examples along with a short explanation of the suggested reasoning, and are encouraged but not required to pick the suggested rating.
For Experiment 1a and 2, where no context is provided, we use the same bigrams, shown in Table \ref{tab:fewshot-no-context}.

\begin{table*}
\centering
    \begin{tabular}{p{0.95\textwidth}}
    \toprule 
    \textbf{Context:} Sarah asks Leo to go to the store to buy a bell pepper. When he gets there, he realizes she didn't say which color pepper he should buy. He buys a green pepper. When he gets home, Sarah is disappointed, because she prefers the red ones. \\
    \textbf{Question:} On a scale of ``Definitely not'', ``Probably not'', ``Unsure'', ``Probably yes'' or ``Definitely yes'', in this context, is the green pepper still a pepper? \\
    \textbf{Answer:} Definitely yes \\
    \\
    \textbf{Context:} Mark is an expert carver and carves a highly realistic pear out of dark colored wood. He hides the wooden pear in his fruit bowl among the fruit he bought from the supermarket. \\
    \textbf{Question:} On a scale of ``Definitely not'', ``Probably not'', ``Unsure'', ``Probably yes'' or ``Definitely yes'', in this context, is the wooden pear still edible? \\
    \textbf{Answer:} Definitely not \\
    \\
    \textbf{Context:} Bob has climbing roses growing all up the side of his house, and wants to trim them for the first time. He needs to find a way to reach the roses higher up. He looks in his shed to see what he has and finds that he has a small ladder, which he can use to reach the roses halfway up the house, though not the ones at the very top. \\
    \textbf{Question:} On a scale of ``Definitely not'', ``Probably not'', ``Unsure'', ``Probably yes'' or ``Definitely yes'', in this context, is the small ladder still useful? \\
    \textbf{Answer:} Unsure \\
    \\
    \textbf{Context:} Sam asks Carla to go to the store to buy ice for drinks for their party. Unfortunately, she leaves it in her car all day and comes back in the evening to find that it has all melted. Carla doesn't know what to say to Sam about the melted ice, which he was planning to use in their cocktails. \\
    \textbf{Question:} On a scale of ``Definitely not'', ``Probably not'', ``Unsure'', ``Probably yes'' or ``Definitely yes'', in this context, is the melted ice still ice? \\
    \textbf{Answer:} Probably not \\
    \\
    \textbf{Context:} Jordan's friend is on the high school basketball team, and is the tallest among her friends. At the match, Jordan notices that her friend is actually a short basketball player, as most of the other players are taller than her.\\
    \textbf{Question:} On a scale of ``Definitely not'', ``Probably not'', ``Unsure'', ``Probably yes'' or ``Definitely yes'', in this context, is the short basketball player still tall? \\
    \textbf{Answer:} Probably yes \\
    \bottomrule
    \end{tabular}
    \caption{Few-shot prompts for Experiment 1b and 3b, which give one bigram/context pair for each value on the Likert scale to demonstrate use of the scale for ``is-a'' judgments.}
    \label{tab:fewshot-contexts}
\end{table*}

\begin{table*}
\centering
    \begin{tabular}{p{0.95\textwidth}}
    \toprule 
    \textbf{Question:} On a scale of ``Definitely not'', ``Probably not'', ``Unsure'', ``Probably yes'' or ``Definitely yes'', is a green pepper still a pepper? \\
    \textbf{Answer:} Definitely yes \\
    \\
    \textbf{Question:} On a scale of ``Definitely not'', ``Probably not'', ``Unsure'', ``Probably yes'' or ``Definitely yes'', is a wooden pear still edible?
    \\
    \textbf{Answer:} Definitely not \\
    \\
    \textbf{Question:} On a scale of ``Definitely not'', ``Probably not'', ``Unsure'', ``Probably yes'' or ``Definitely yes'', is a small ladder still useful?
    \\
    \textbf{Answer:} Unsure \\
    \\
    \textbf{Question:} On a scale of ``Definitely not'', ``Probably not'', ``Unsure'', ``Probably yes'' or ``Definitely yes'', is melted ice still ice?
    \\
    \textbf{Answer:} Probably not \\
    \\
    \textbf{Question:} On a scale of ``Definitely not'', ``Probably not'', ``Unsure'', ``Probably yes'' or ``Definitely yes'', is a short basketball player still tall? \\
    \textbf{Answer:} Probably yes \\
    \bottomrule
    \end{tabular}
    \caption{Few-shot prompts for Experiment 1a and Experiment 2, which give one bigram for each value on the Likert scale to demonstrate use of the scale for ``is-a'' judgments.}
    \label{tab:fewshot-no-context}
\end{table*}

\subsection{Prompts for context generation} \label{app:generation-prompts}

We use the sequence of prompts shown in Table \ref{tab:context-gen-prompts} to generate 12 contexts (``stories'') in a single chat. We use three contexts already used in the few-shot prompts for Experiment 1 as example stories to help control the style and level of complexity of the language. We found that providing example stories was much more effective than trying to control the output with detailed instructions. \{bigram\} and \{a/an\} are substituted in at runtime.

\begin{table*}
\centering
    \begin{tabular}{p{0.95\textwidth}}
    \toprule 
    In this task, you will write short, simple stories of 50-100 words about specific objects or things. The story should use simple language and describe the object in detail. \\
    \\
    \textbf{Example:} Write a short, simple story about a wooden pear.\\
    \textbf{Story:} Mark is an expert carver and carves a highly realistic pear out of dark colored wood. He hides the wooden pear in his fruit bowl among the fruit he bought from the supermarket. When his friends come to visit, only one of them noticed the wooden pear hiding among the fruit. \\
    \\
    \textbf{Example:} Write a short, simple story about melted ice. \\
    \textbf{Story:} Sam asks Carla to go to the store to buy ice for drinks for their party. Unfortunately, she leaves it in her car all day and comes back in the evening to find that it has all melted. Carla doesn't know what to say to Sam about the melted ice, which he was planning to use in their cocktails. \\
    \\
    \textbf{Example:} Write a short, simple story about a short basketball player. \\
    \textbf{Story:} Jordan's friend is on the high school basketball team, and is the tallest among her friends. At the match, Jordan notices that her friend is actually a short basketball player, as most of the other players are taller than her. Even so, her friend does very well in the game and scores six points. \\
    \\
    Now, write a short, simple story about \{a/an\} \{bigram\}, which includes the phrase ``\{bigram\}'' and describes the {bigram} in detail. Start your story with ``Story:''. \\ \midrule
    Write another different story about \{a/an\} \{bigram\}. Start your story with ``Story:''. \textit{(repeated 2x)} \\ \midrule
    Great! Write three more stories about \{a/an\} \{bigram\}. Number your stories with ``Story 1:'', ``Story 2:'' and ``Story 3:''. \textit{(repeated 3x)} \\
    \bottomrule
    \end{tabular}
    \caption{Prompts for Experiment 3a to generate contexts featuring a given  bigram in which the inference ``Is an \{adjective\} \{noun\} a \{noun\}?'' will later be judged.}
    \label{tab:context-gen-prompts}
\end{table*}

\section{Additional qualitative analysis} 

\subsection{Context generation}

In addition to the regression in Section \ref{sec:method-comparison}, we conduct a qualitative error analysis of the bigrams where the context generation method's JS divergence from the human distribution is particularly high.
Manual inspection of these bigrams suggests that some issues likely stem from Llama 3 70B Instruct not picking same the (literal or abstract) noun meaning as humans: for example, most contexts for \textit{useful heart} involve a metaphorical \textit{heart} (e.g.~the heart of a community), while \textit{false market} contexts all involve physical markets rather than economic markets. 
In some cases, the model also appears to interpret the adjective metaphorically or loosely: judging by the generated contexts, Llama 3 seems to think that a \textit{fake idea} is simply a \textit{wrong idea} or \textit{false idea}. 
While \textit{fake} can sometimes mean \textit{false} and vice versa, this is not the intuition the authors had about \textit{fake idea}.
Finally, some bigrams also get misunderstood as noun-noun compounds such as \textit{counterfeit attack} in the generated context in Appendix \ref{app:generated-contexts}, or incorporated into longer noun-noun compounds, such as \textit{useful attack dog} or \textit{former attack helicopter}. 

More broadly, while the LLM generates pleasingly diverse contexts for examples like \textit{fake crowd}, other examples such as \textit{fake concert} are lacking in diversity, with essentially the same blueprint being repeated in all 12 stories.

\section{Generated contexts} \label{app:generated-contexts}

To illustrate the qualitative analysis in Section \ref{sec:method-comparison}, we provide a representative context for \textit{homemade cat} and two for \textit{fake leg}.
We further show 9 randomly sampled contexts generated by Llama 3 70B Instruct, one for each of the 6 typically-privative adjectives, and one context generated for the typically-subsective adjectives \textit{illegal}, \textit{useful} and \textit{multicolored}. (Contexts for the typically-subsective adjectives tend to follow an uninteresting, predictable pattern; we stumbled across this interesting supposedly-privative example involving \textit{multicolored}).

\begin{enumerate}
    \item \textbf{Context:} Ben's mom is a talented seamstress, and she helps him make a \textbf{homemade cat} out of old t-shirts and fabric remnants. They stuff the cat with polyester fiberfill and add a sweet face made from felt. Ben names the cat Stitch and takes it to school for show-and-tell. His classmates ooh and ahh over Stitch's colorful patches and softness. \\
    \textbf{LLM Rating:} Probably not (privative)
    \item \textbf{Context:} In the museum of medical history, a \textbf{fake leg} was on display as part of an exhibit on the evolution of prosthetics. The fake leg was made in the 19th century, and was crafted from wood and leather. It was a crude but functional device, with a simple hinge at the knee and a leather strap to hold it in place. Visitors marveled at the primitive technology, and were amazed by how far prosthetic design had come since then. The fake leg was a testament to the ingenuity and perseverance of people with disabilities throughout history. \\
    \textbf{LLM Rating:} Probably not (privative)
    \item \textbf{Context:} Rachel, a young amputee, was thrilled to receive her new \textbf{fake leg} from the hospital. It was her first prosthetic, and she couldn't wait to start learning how to walk again. The fake leg was custom-made to fit her perfectly, with a soft, cushioned socket and a sturdy metal limb. With the help of her physical therapist, Rachel quickly learned to navigate the fake leg, and soon she was running and playing with her friends like she used to. \\
    \textbf{LLM Rating:} Probably yes (subsective)
    \item \textbf{Context:} At the sci-fi convention, attendees are amazed by an \textbf{artificial painting} that appears to come to life. The painting, created by a team of digital artists, depicts a fantastical landscape with towering mountains and glowing forests. As people walk by, the painting responds by changing color, shape, and even emitting sounds and scents. The artificial painting is a huge hit, and people line up to experience its immersive, interactive magic. \\
    \textbf{LLM Rating:} Definitely yes (subsective)
    \item \textbf{Context:} The detective was called to investigate a \textbf{counterfeit attack} on a popular online shopping website. The hackers had created fake websites that looked identical to the real one, tricking customers into entering their credit card information. The detective worked tirelessly to track down the culprits, following a trail of digital clues that led him from server to server. Finally, he identified the IP address of the hackers and worked with the authorities to shut down the operation, saving thousands of people from financial loss. \\
    \textbf{LLM Rating:} Definitely yes (subsective) \\
    \textbf{Authors' note: } The LLM interpreted \textit{counterfeit attack} as a noun-noun compound here, not as an adjective-noun combiniation (i.e. it is an attack involving counterfeits, not an attack that is counterfeit).
    \item \textbf{Context:} Rachel is a set designer for a film production company. She needs a \textbf{fake stick} that looks like a real magic wand for an upcoming fantasy movie. She searches high and low and finally finds the perfect fake stick, complete with intricate carvings and a sparkling crystal at the tip. The fake stick becomes a key prop in the movie, and the actors love using it during filming. The movie goes on to become a huge hit, and Rachel's fake stick is praised as one of the most iconic props in the film. \\
    \textbf{LLM Rating:} Definitely yes (subsective) 
    \item \textbf{Context:} The antique shop owner was thrilled to find an old chest in the attic of his newly inherited store. As he rummaged through it, he found a stack of old bills that looked like they dated back to the 1800s. But as he examined them closer, he realized they were actually \textbf{false currency}, likely created by a counterfeiter from that era. Despite being worthless, the bills were still a fascinating piece of history, and the owner decided to display them in his shop as a curiosity. \\
    \textbf{LLM Rating:} Probably not (privative)
    \item \textbf{Context:} Jack's \textbf{former reason} for going to the beach every summer was a faded, beach towel that his family had used for years. The towel had a fun, colorful design and a soft, worn-out texture, but it was more than just a towel - it was a symbol of family vacations and summer fun. Jack would wrap himself in it after a long day of swimming, but after his family stopped going to the beach, the towel was forgotten in the back of a closet, a reminder of his former reason for soaking up the sun. \\
    \textbf{LLM Rating:} Definitely not (privative) 
    \item \textbf{Context:} Lily's friend Emma lends her a \textbf{knockoff jewel} to wear to a formal dance. The knockoff jewel is a beautiful, long necklace with a big, fake pearl pendant. Lily loves the necklace and feels like a princess when she wears it. However, when she tries to return it to Emma, she realizes that she lost the clasp and can't find it anywhere. Emma is upset, but Lily promises to buy her a new one to replace it. \\
    \textbf{LLM Rating:} Definitely yes (subsective) 
    \item \textbf{Context:} In the small village of Ravenswood, a mysterious \textbf{illegal currency} began to circulate. The coins were made of a heavy, dark metal and featured an image of a raven on one side and a crescent moon on the other. They were stamped with a strange symbol that no one could decipher. At first, the villagers were wary of the illegal currency, but as more and more merchants began to accept it, it became a popular alternative to the official coins. However, the village elder warned that the authorities would soon crack down on the illegal currency, and those caught using it would face severe penalties. \\
    \textbf{LLM Rating:} Definitely yes (subsective)
    \item \textbf{Context:} In a small village, there lived a respected elder named Nalani who used a \textbf{useful fruit} - a sweet and nutritious mango with a yellow skin and creamy flesh. Nalani would dry mango slices to create a natural sweetener for her traditional recipes, and she would also use mango oil to nourish her skin and hair. The villagers would come to Nalani for beauty tips, and she would share her secrets of using the mango to promote healthy skin and hair. Nalani's wisdom and knowledge of the useful mango were cherished by the community, and she was revered as a guardian of traditional knowledge. \\
    \textbf{LLM Rating:} Definitely yes (subsective) 
    \item \textbf{Context:} On the beach, Jack builds an intricate sandcastle, topped with a \textbf{multicolored image} of a dragon. The dragon's scales glint in the sunlight, made from tiny shells and pebbles in shades of green, blue, and purple. As the tide comes in, Jack watches as the dragon begins to dissolve, its multicolored image slowly disappearing into the waves. He smiles, knowing that the memory of his creation will stay with him long after the castle is gone. \\
    \textbf{LLM Rating:} Probably not (privative)
\end{enumerate}

\section{Additional experiment results} \label{app:result-figures-tables}

\begin{table}[t]
    \centering
    \begin{tabular}{lcc}
    \toprule
    JS divergence & Privative & Subsective \\ \midrule
    Both low ($<0.25$) & 103 & 339  \\
    Both high ($>0.5$) & 31 & 3 \\
    Log-prob $>$ context & 187 & 39 \\
    Context $>$ log-prob & 45 & 7 \\
    \bottomrule
    \end{tabular}
    \caption{Number of bigrams with privative vs.~subsective adjectives where Jensen-Shannon divergence is low (<0.25) for both Method 2 (log-probability) and Method 3 (context generation), high (>0.5) for both, or where one Method is (<0.5 and) better than the other.}
    \label{tab:method-comparison-bigram-counts}
\end{table}

\subsection{Tabular numerical results}

Detailed numerical results for the figures in the main body of the paper, including results for the base models, are shown in Tables \ref{tab:logprob-js-base}, \ref{tab:exp-1b-context}, \ref{tab:exp-1b-context-1sd} and \ref{tab:exp2-1sd}. 
Table \ref{tab:method-comparison-bigram-counts} shows thresholded counts for how many bigrams each method of obtaining a distribution performs better for.

\subsection{Accuracy within 1 SD for Experiment 1b}

Figure \ref{fig:scaling-base-1sd} shows Experiment 1b, which predicts the inference given the context, using the more lenient accuracy within 1 SD of the human mean metric introduced in Section \ref{sec:acc-1-sd-method}. 
This more lenient metric does not penalise models which use ``Unsure'', provided that that is within human ratings.
With this metric, performance scales uniformly with size for instruction-tuned models.

\subsection{Experiment 1b: Base models}

\begin{figure*}[t]
  \includegraphics[width=\textwidth]{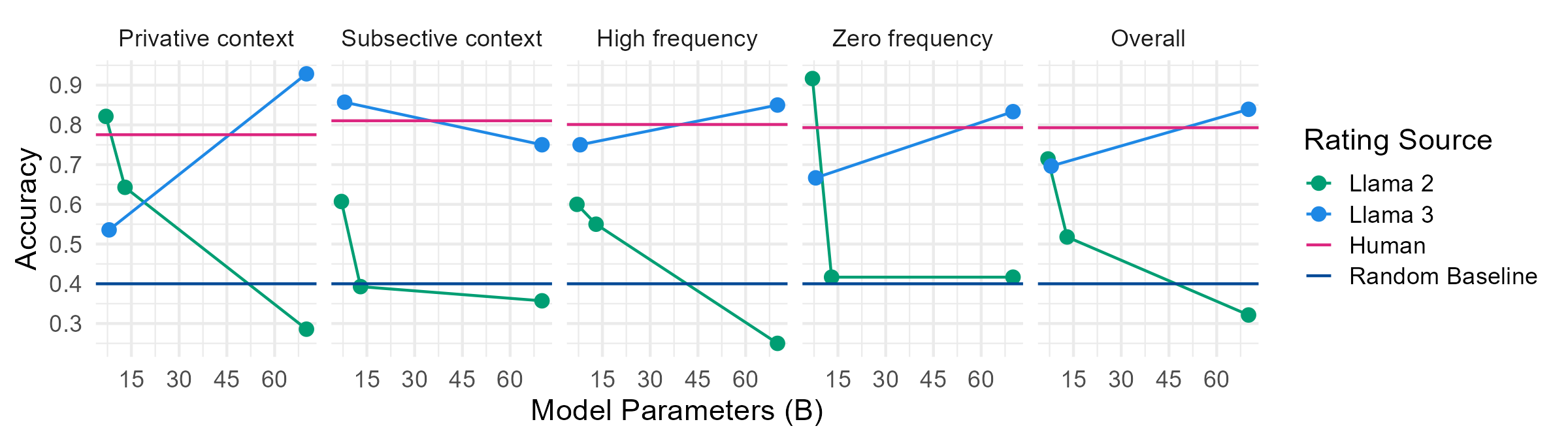}
  \caption{Accuracy on the context-based inference task for base models (5-shot). Accuracy increases with parameters for Llama 3, but drops for Llama 2. However, accuracy within 1 SD on this task increases with model parameters -- see Figure \ref{fig:scaling-base-1sd}.
  }
  \label{fig:scaling-base}
\end{figure*}

\begin{figure*}[t]
  \includegraphics[width=\textwidth]{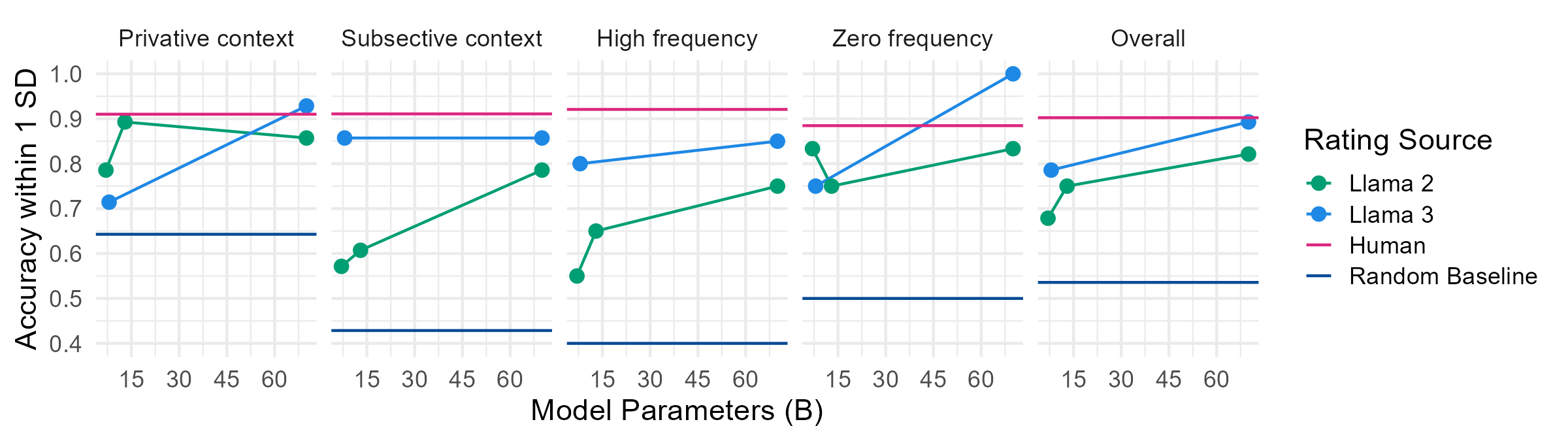}
  \caption{Accuracy within 1 SD of the human mean on the context-based inference task for base models (5-shot). Under this more lenient metric, overall accuracy increases with model size, though still not in every category for Llama 2.
  }
  \label{fig:scaling-base-1sd}
\end{figure*}

For the inference task where the context is provided, Experiment 1b, we see in Figure \ref{fig:scaling-base} and Table \ref{tab:exp-1b-context} that Llama 3 70B actually out-performs its Instruct model overall by 7 points of accuracy. This is largely due to its high accuracy on privative contexts (0.93 instead of 0.61), which compensates for its lower accuracy on subsective contexts (0.75 instead of 0.93). 
We see the same pattern for Llama 3 8B, which is better at privative contexts and worse at subsective contexts than its instruction-tuned counterpart. 
One possibility is that Llama 3 Instruct is generally more biased to affirmative ratings (i.e.~subsective ratings on this scale) after instruction/helpfulness tuning, thus doing better in the subsective context simply because a subsective rating is correct there. 
We can see in Table \ref{tab:exp1a-no-context} that when no context is provided (Experiment 1a), Llama 3 Instruct assigns subsective and privative ratings roughly evenly across bigrams, while Llama 3 has a bias (69.2\%) towards giving these bigrams with typically-privative adjectives privative ratings.
For Llama 3, we see scaling with size for both the accuracy metric and the accuracy within 1 SD metric, as shown in Figure \ref{fig:scaling-base} and Figure \ref{fig:scaling-base-1sd}.

This pattern does not occur in Llama 2 across the board - we see a marked improvement for Llama 2 7B over Llama 2 7B Chat in both context types, but Llama 2 13B Chat is better at privative contexts than Llama 2 13B and only slightly worse at subsective contexts, suggesting no overall pattern for Llama 2, and no effect of its type of instruction tuning. 
Likewise in Table \ref{tab:exp1a-no-context}, we see that instruction-tuning Llama 2 70B does not result in a large shift in the ratio of privative to subsective ratings, but rather mostly reduces the proportion of ``Unsure'' ratings.
Notably, we actually see inverse scaling with size for Llama 2 on all splits of this task (see Figure \ref{fig:scaling-base}) when using the accuracy metric, but regular scaling with size using the accuracy within 1 SD metric, which does not punish the ``Unsure'' rating so harshly (Figure \ref{fig:scaling-base-1sd}).

\subsection{Experiment 2: Base models}

\begin{figure*}[t]
  \includegraphics[width=\textwidth]{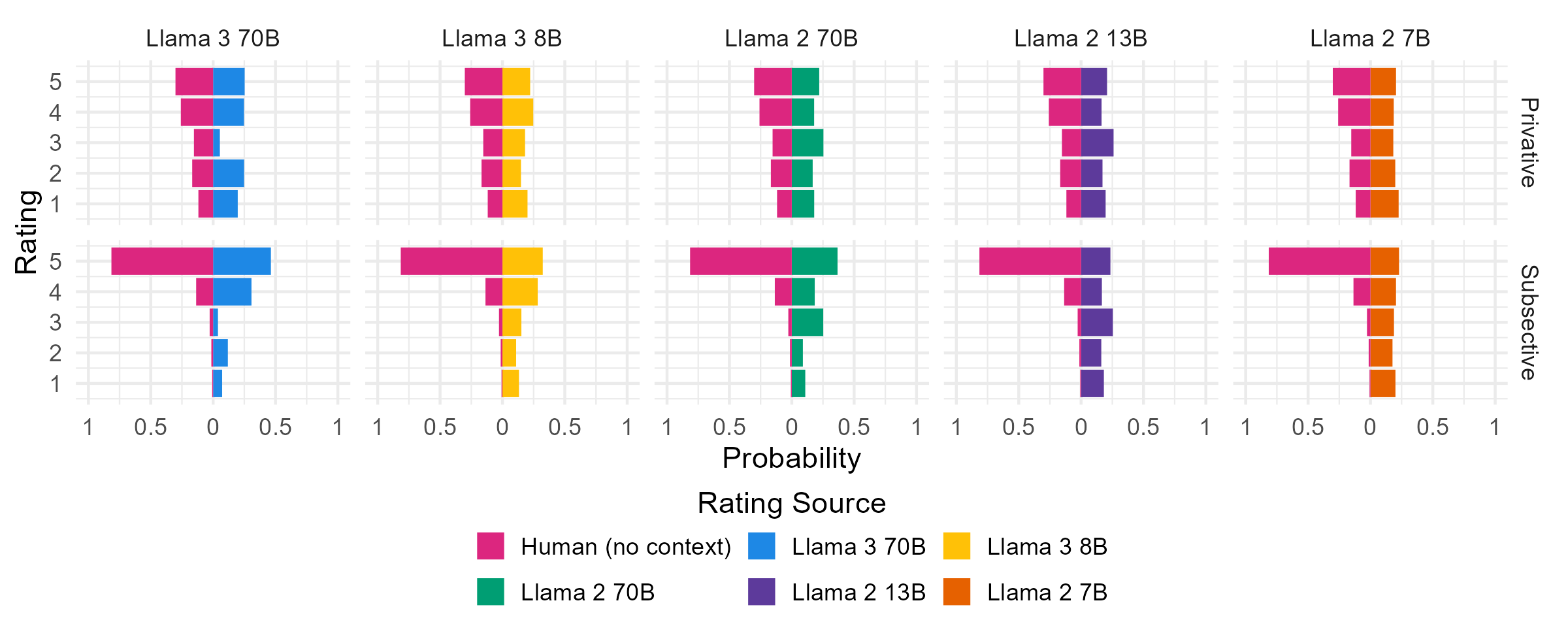}
  \caption{Average log-probability distribution for (typically) subsective vs.~privative adjectives for base LLMs, compared to the average human distribution.}
  \label{fig:logprob-distributions-by-adjclass-base}
\end{figure*}

On the inference task where no context is provided, Experiment 2, we see in Table \ref{tab:exp2-1sd} that Llama 3 70B, Llama 3 8B and Llama 2 70B perform comparably to their instruction-tuned counterparts using the within 1 SD metric. 
Performance is less predictable for the smaller Llama 2 models, with the ranking Llama 2 13B Chat > Llama 2 7B > Llama 2 13B > Llama 2 7B Chat. 
As discussed in the main body of the paper, this metric is quite lenient -- the fact that Llama 2 7B Chat actually \emph{underperforms} the random baseline, and that the smaller Llama 2 models underperform the ``majority'' baseline by at least 20 points of accuracy), is itself striking, suggesting a distinct lack of comprehension of the task when presented out of the blue with no context. 
While some of this difficulty may be attributed to the Likert scale, even these small models perform at at least 60\% accuracy using the same metric and Likert scale (Table \ref{tab:exp-1b-context-1sd}) when a context is provided, so the difficulty must lie at least partially with the out-of-the-blue setting.
Perhaps these older, smaller models do not capture enough information about how the world typically is from their pretraining (do not have sufficiently human-like ``priors'' or world knowledge). 
Looking at the distributions they produce, Figure \ref{fig:logprob-distributions-by-adjclass-base} shows that they distribute their probability mass relatively evenly across the scale for subsective adjectives, resulting in a poor fit. Table \ref{tab:logprob-js-base} shows the Jensen-Shannon divergences, which are relatively low for subsective adjectives for these models. 
Interestingly, however, models of all sizes are competitive for fitting the distribution of typically-privative adjectives overall, with Qwen 2 72B Instruct, Mixtral 7x8B Instruct and Llama 2 7B Chat scoring the lowest at around 0.3 JS divergence. Fitting the human distribution of privative adjectives is not a function of model size at all.

\begin{table}[t]
    \centering
    \begin{tabular}{lccc}
    \toprule
    & \multicolumn{3}{c}{JS Divergence} \\
    Model & Priv. & Subs. & Total \\ \midrule
    Human & 0 & 0 & 0 \\
    Llama 3 70B Instruct & 0.26 & 0.08 & 0.17 \\
    Qwen 2 72B Instruct & 0.33 & 0.08 & 0.19  \\
    Llama 3 70B & 0.16 & 0.21  & 0.19   \\
    Llama 2 70B Chat & 0.18 & 0.25 & 0.22 \\
    Mixtral 7x8B Instruct & 0.32 & 0.13 & 0.22 \\
    Llama 2 70B & 0.17 & 0.30  & 0.24 \\
    Llama 3 8B & 0.18 & 0.32 & 0.26   \\
    Llama 3 8B Instruct & 0.18 & 0.34 & 0.26 \\
    Llama 2 13B Chat  & 0.25 & 0.35 & 0.30 \\
    Llama 2 7B & 0.20 & 0.43  & 0.32 \\ 
    Llama 2 13B & 0.21 & 0.43 & 0.32  \\
    Uniform baseline & 0.20 & 0.46 & 0.34 \\
    Llama 2 7B Chat  & 0.29 & 0.46 & 0.38 \\
    ``Majority'' baseline & 0.71 & 0.12 & 0.40 \\
    \bottomrule
    \end{tabular}
    \caption{Jensen-Shannon divergence between per-bigram rating distributions for humans and LLMs when sourced from log-probabilities, for privative vs.~subsective adjectives, including base models.}
    \label{tab:logprob-js-base}
\end{table}

\subsection{Experiment 1c: Zero-shot inferences with context} \label{app:zero-shot}

Experiment 1c performs an ablation study on Experiment 1b where we run the same experiment of determining the inference given a biasing context, but 0-shot instead of with 5-shot examples of ``is-a'' inferences on the Likert scale.
Table \ref{tab:exp-1c-context-0shot} shows the results of Experiment 1c. 

Whether the 5-shot examples help or hinder depend on the individual model. For Llama 3 70B Instruct, Llama 2 13B Chat, Llama 2 7B Chat and Mixtral 7x8B Instruct, we see a drop of 3-14 points in accuracy when we prompt the model 0-shot.
For the other three instruct models, however, Llama 2 70B Chat, Llama 3 8B Instruct and Qwen 2 72B Instruct, we see a 6-9 point increase in accuracy when we prompt 0-shot, suggesting that these models found the 5 examples (which were examples of using the scale with ``is-a'' inferences, but were not exactly the target task which always repeated the noun) misleading rather than helpful.

Turning to the base models, we see a 4-13 point drop for all models (Llama 2 70B, Llama 3 8B, Llama 2 13B, Llama 2 7B) except Llama 2 70B when prompted 0-shot instead of 5-shot. Llama 2 70B shows a 38 point \emph{increase} in accuracy on this task when prompted 0-shot. This is because its very low score on the 5-shot task is largely caused by it answering ``Unsure'' for many bigrams, which is always considered incorrect under this metric. Without an example where ``Unsure'' is used, even though ``Unsure'' is mentioned as a scale item, Llama 2 70B uses this rating far less often and is able to score much higher on this task.
This is in part an artefact of how this metric is scored -- recall from Figure \ref{fig:scaling-base-1sd} that Llama 2 70B still scores well on the more lenient accuracy within 1 SD metric, which includes ``Unsure'' for many bigrams.

\subsection{Regression details for Section \ref{sec:method-comparison}}

\begin{figure*}[t]
\centering
\begin{subfigure}{0.49\textwidth}
   \includegraphics[width=1\linewidth]{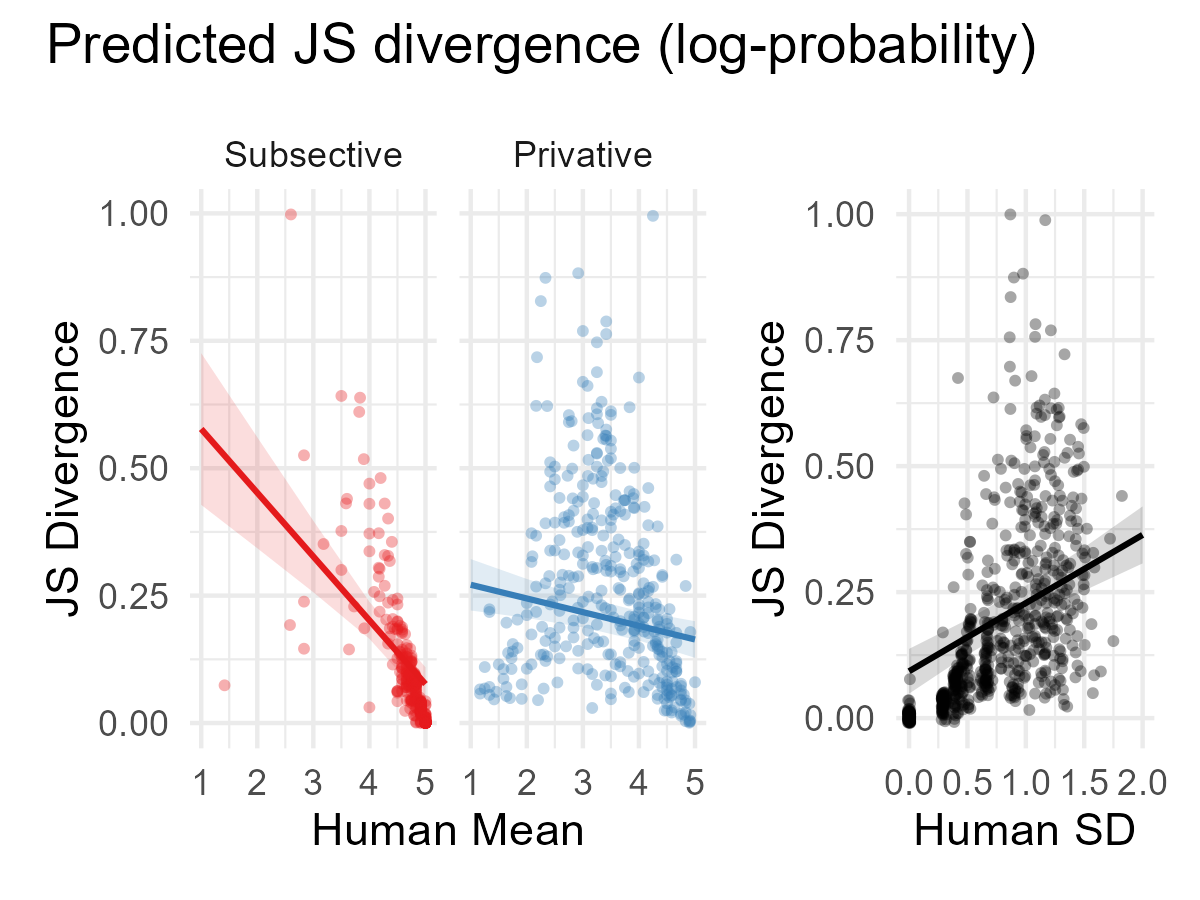}
   \caption{Log-probability}
   \label{fig:logprob-method-regression-effects} 
\end{subfigure}
\begin{subfigure}{0.49\textwidth}
   \includegraphics[width=1\linewidth]{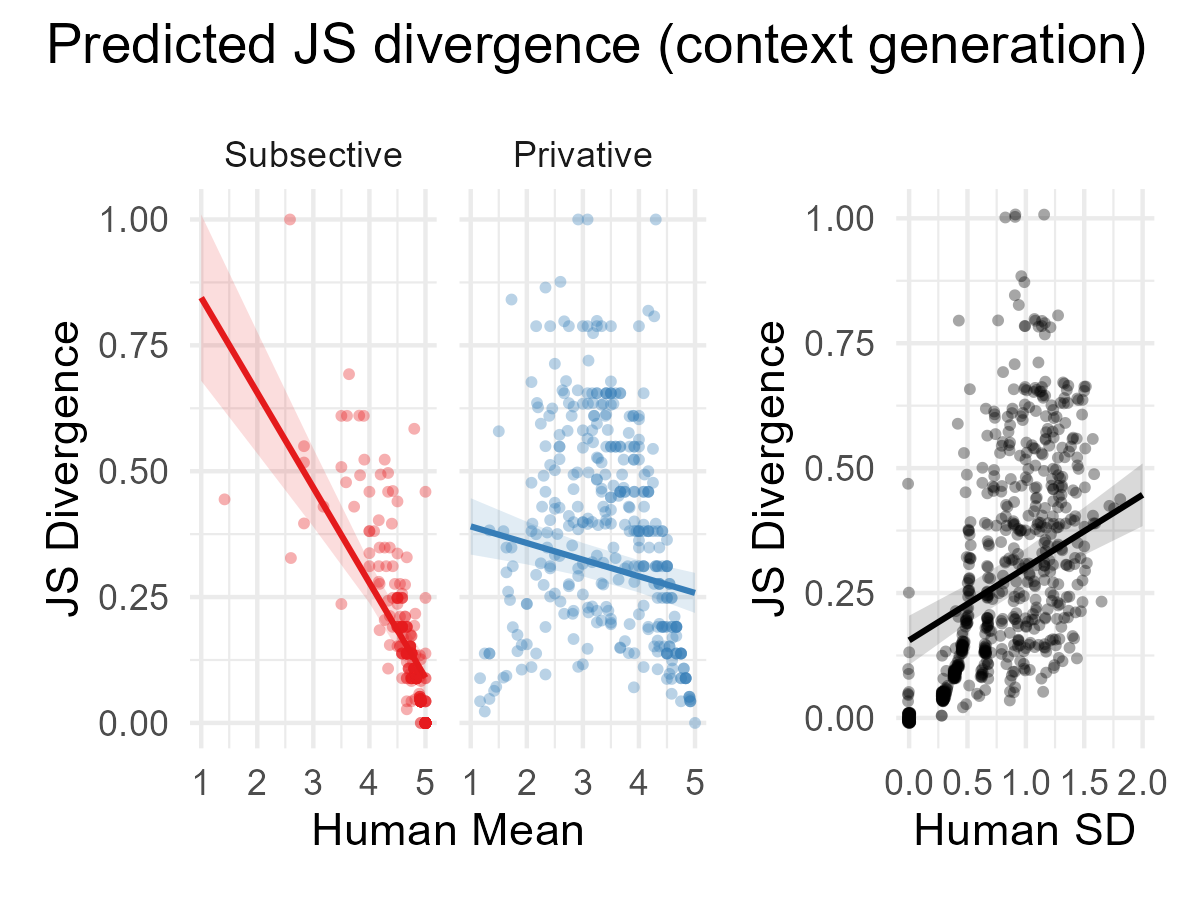}
   \caption{Context generation}
   \label{fig:context-method-regression-effects}
\end{subfigure}
\caption{Effects plots for the regressions \texttt{JSDivergence $\sim$ AdjectiveType * HumanMean + HumanSD + BigramFrequency} for each method in Section \ref{sec:method-comparison}. There is no significant effect of bigram frequency.}
\label{fig:method-regression-effects}
\end{figure*}

Table \ref{tab:method-regression} and Figure \ref{fig:method-regression-effects} show the coefficients and effects plots for the two regressions in Section \ref{sec:method-comparison}.

\begin{table}[t]
    \centering
    \begin{tabular}{lcc}
    \toprule
    Coefficient & $\hat{\beta}$ & $p$   \\ \midrule
    \textit{\textbf{Log-probability}} && \\
    Intercept & 0.61 & $p < 0.01$ \\
    Privative & -0.40 & $p < 0.01$ \\
    Human mean & -0.13 & $p < 0.01$ \\
    Human SD & 0.14 & $p < 0.01$ \\
    75\textsuperscript{th}-90\textsuperscript{th} percentile & 0.01 & $p = 0.56$ \\
    50\textsuperscript{th}-75\textsuperscript{th} percentile & 0.02 & $p = 0.22$ \\
    25\textsuperscript{th}-50\textsuperscript{th} percentile & 0.02 & $p = 0.25$ \\
    Zero frequency & 0.01 & $p = 0.36$ \\
    Privative:Human mean & 0.10 & $p < 0.01$ \\ \midrule
    \textit{\textbf{Context generation}} && \\
    Intercept & 0.93 & $p < 0.01$ \\
    Privative & -0.61 & $p < 0.01$ \\
    Human mean & -0.19 & $p < 0.01$ \\
    Human SD & 0.15 & $p < 0.01$ \\
    75\textsuperscript{th}-90\textsuperscript{th} percentile & 0.02 & $p = 0.38$ \\
    50\textsuperscript{th}-75\textsuperscript{th} percentile & 0.03 & $p = 0.15$ \\
    25\textsuperscript{th}-50\textsuperscript{th} percentile & 0.03 & $p = 0.16$ \\
    Zero frequency & 0.03 & $p = 0.09$ \\
    Privative:Human mean & 0.16 & $p < 0.01$ \\
    
    \bottomrule
    \end{tabular}
    \caption{Coefficients for the regressions \texttt{JSDivergence $\sim$ AdjectiveType * HumanMean + HumanSD + BigramFrequency} for each method in Section \ref{sec:method-comparison}. For the adjective type factor, subsective is the first level (intercept), for frequency bins, 90th-99th percentile is the first level (intercept); dummy coding is used throughout.}
    \label{tab:method-regression}
\end{table}

\begin{table*}[t]
    \centering
    \begin{tabular}{lccccc}
    \toprule
    & \multicolumn{5}{c}{Accuracy} \\
    Model & Privative context & Subsective context & High freq. & Zero freq. & Total \\ \midrule
    Human & 0.78 & 0.81 & 0.80 & 0.79 & 0.79 \\
    Llama 3 70B & 0.93 & 0.75 & 0.85 & 0.83 & 0.84 \\
    Qwen 2 72B Instruct & 0.68 & 0.93 & 0.80 & 0.83 & 0.80 \\
    Llama 3 70B Instruct  & 0.61 & 0.93 & 0.80 & 0.75 & 0.77 \\
    Mixtral 7x8B Instruct & 0.79 & 0.68 & 0.80 & 0.75 & 0.73 \\
    Llama 2 7B & 0.82 & 0.61 & 0.60 & 0.92 & 0.71 \\ 
    Llama 3 8B & 0.54 & 0.86  & 0.75 & 0.67 & 0.70  \\
    Llama 3 8B Instruct & 0.29 & 0.96 & 0.60 & 0.58 & 0.63 \\
    Llama 2 13B Chat & 0.89 & 0.36 & 0.60 & 0.67 & 0.63 \\
    Llama 2 70B Chat & 0.50 & 0.68 & 0.70 & 0.42 & 0.59 \\
    Llama 2 13B & 0.64 & 0.39  & 0.55 & 0.42 & 0.52  \\
    Llama 2 7B Chat & 0.75 & 0.25 & 0.50 & 0.50 & 0.50 \\ 
    Llama 2 70B & 0.29 & 0.36  & 0.25 & 0.42 & 0.32 \\
    Random baseline & 0.4 & 0.4 & 0.4 & 0.4 & 0.4 \\
    \bottomrule
    \end{tabular}
    \caption{Accuracy on the (5-shot) context-based inference task (Experiment 1) overall, by bigram frequency and by context bias. 
    }
    \label{tab:exp-1b-context}
\end{table*}

\begin{table*}[t]
    \centering
    \begin{tabular}{lccccc}
    \toprule
    & \multicolumn{5}{c}{Accuracy} \\
    Model & Privative context & Subsective context & High freq. & Zero freq. & Total \\ \midrule
    Human & 0.91 & 0.91 & 0.92 & 0.89 & 0.90 \\
    Llama 3 70B & 0.93 & 0.86 & 0.85 & 1.00  & 0.89 \\
    Llama 2 70B Chat & 0.89 & 0.86 & 0.85 & 0.83 & 0.88 \\
    Qwen 2 72B Instruct & 0.79 & 0.93 & 0.85 & 0.83 & 0.86 \\
    Llama 2 70B & 0.86 & 0.79 & 0.75 & 0.83 & 0.82 \\
    Llama 3 70B Instruct & 0.61 & 1.00 & 0.80 & 0.92  & 0.80 \\
    Llama 3 8B & 0.71 & 0.86 & 0.80 & 0.75 & 0.79 \\
    Llama 2 13B & 0.89 & 0.61 & 0.65 & 0.75 & 0.75 \\
    Mixtral 7x8B Instruct & 0.75 & 0.64 & 0.70 & 0.67 & 0.70 \\
    Llama 3 8B Instruct & 0.46 & 0.89 & 0.65 & 0.67 & 0.68  \\
    Llama 2 7B Chat & 0.79 & 0.57 & 0.55 & 0.83 & 0.68  \\
    Llama 2 7B & 0.79 & 0.57 & 0.55 & 0.83  & 0.68  \\
    Llama 2 13B Chat & 0.93 & 0.36 & 0.60 & 0.75 & 0.64 \\
    Random baseline & 0.64 & 0.43 & 0.40 & 0.50  & 0.54  \\
    \bottomrule
    \end{tabular}
    \caption{Accuracy within 1 SD of human mean on the (5-shot) context-based inference task (Experiment 1) overall, by bigram frequency and by context bias. 
    }
    \label{tab:exp-1b-context-1sd}
\end{table*}

\begin{table*}[t]
    \centering
    \begin{tabular}{lccccc}
    \toprule
    & \multicolumn{5}{c}{Accuracy ($\mu \pm 1\sigma$)} \\
    Model & Privative & Subsective & High frequency & Zero frequency & Total \\ \midrule
    Human & 0.903 & 0.960 & 0.939 & 0.933 & 0.933 \\
    Qwen 2 72B Instruct & 0.886 & 0.995 & 0.946 & 0.967 & 0.944 \\
    Llama 3 70B Instruct  & 0.778 & 0.995 & 0.892 & 0.911 & 0.892 \\
    Llama 3 70B & 0.815 & 0.960  & 0.907 & 0.928 & 0.891 \\
    ``Majority'' baseline & 0.781 & 0.993 & 0.881 & 0.912 & 0.885 \\
    Llama 2 70B & 0.870 & 0.886 & 0.882 & 0.850 & 0.878 \\
    Llama 2 70B Chat  & 0.831 & 0.767 & 0.824 & 0.722 & 0.797 \\
    Mixtral 7x8B Instruct & 0.653 & 0.914 & 0.803 & 0.756 & 0.791 \\
    Llama 3 8B & 0.765 & 0.795  & 0.806  & 0.783 & 0.781 \\
    Llama 3 8B Instruct & 0.833 & 0.679 & 0.735 & 0.789 & 0.752  \\
    Llama 2 13B Chat  & 0.455 & 0.776 & 0.706 & 0.517 & 0.624 \\
    Llama 2 7B & 0.336 & 0.807 & 0.649 & 0.472 & 0.584 \\ 
    Analogy baseline & 0.648 & 0.431 & 0.527 & 0.567 & 0.534 \\
     Llama 2 13B & 0.772 & 0.317  & 0.566  & 0.433  & 0.533 \\
    Random baseline & 0.610 & 0.325 & 0.464 & 0.456 & 0.460 \\
    Llama 2 7B Chat  & 0.447 & 0.252  & 0.394 & 0.267 & 0.345  \\ 
    \bottomrule
    \end{tabular}
    \caption{Accuracy within 1 SD of the human mean on the (5-shot) no-context inference task (Experiment 2) by bigram frequency and by adjective type (typically-privative or typically-subsective).}
    \label{tab:exp2-1sd}
\end{table*}

\begin{figure*}[t]
  \includegraphics[width=\textwidth]{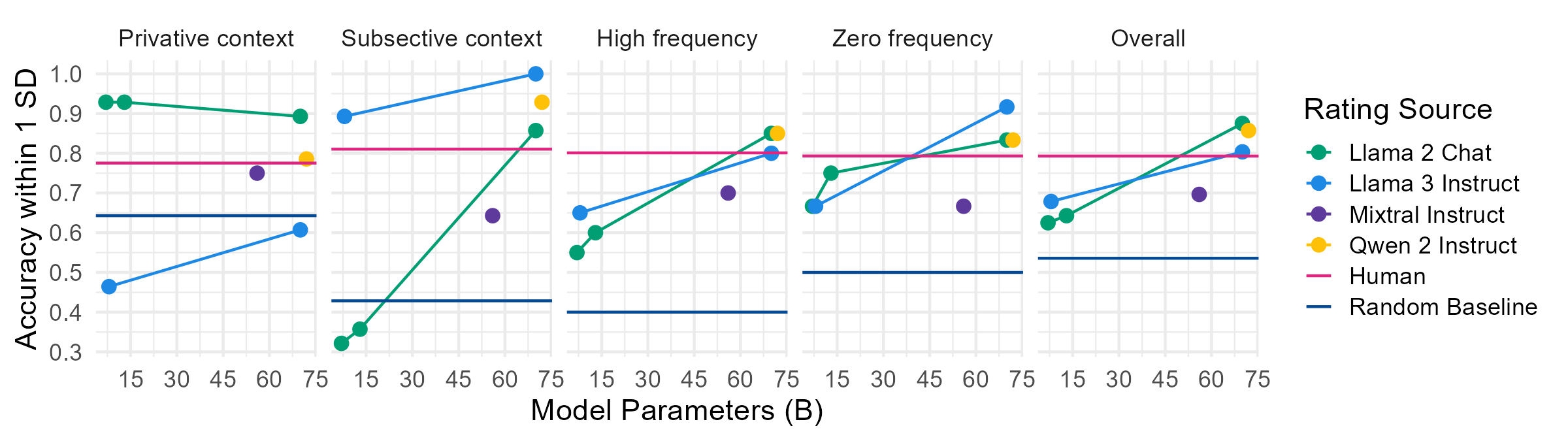}
  \caption{Accuracy within 1 SD of the human mean on the context-based inference task for instruction-tuned models (5-shot). Under this more lenient metric, accuracy increases with model size for all models.
  }
  \label{fig:scaling-instruct-1sd}
\end{figure*}

\begin{figure*}[t]
  \includegraphics[width=\textwidth]{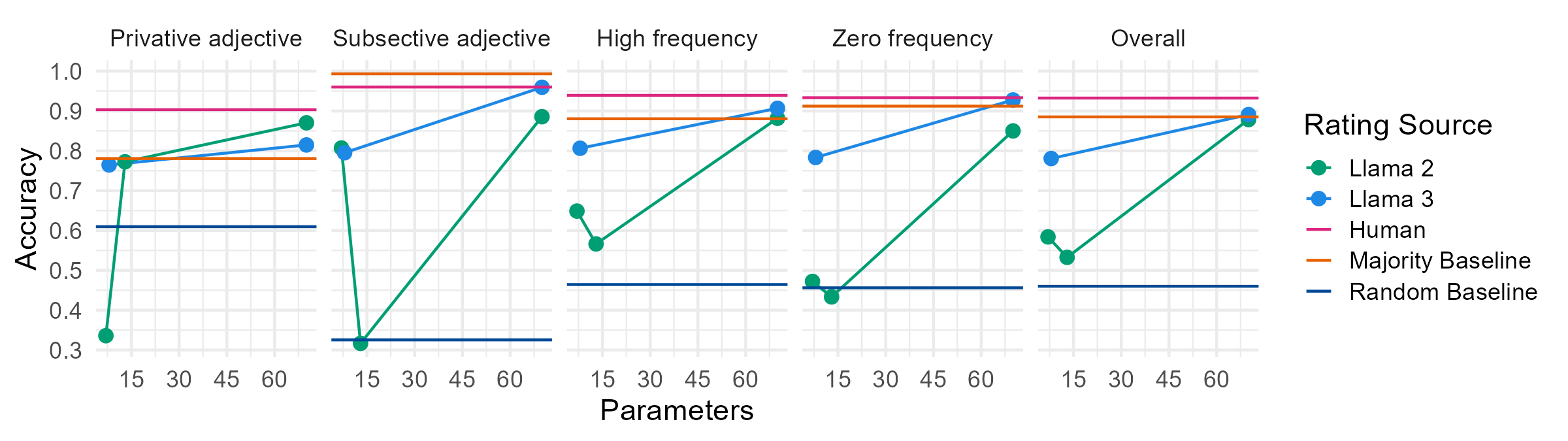}
  \caption{Accuracy within 1 SD of the human mean on the no-context inference task (Experiment 2) by bigram frequency and by adjective type (typically-privative or typically-subsective) for base models (5-shot).}
  \label{fig:contextfree-1sd-scaling-base}
\end{figure*}

\begin{table*}[t]
    \centering
    \begin{tabular}{lccccc}
    \toprule
    & \multicolumn{5}{c}{Accuracy} \\
    Model & Privative context & Subsective context & High freq. & Zero freq. & Total \\ \midrule
    Human & 0.78 & 0.81 & 0.80 & 0.79 & 0.79 \\
    Qwen 2 72B Instruct & 0.79 & 0.93 & 0.85 & 0.83 & 0.86 \\
    Llama 3 70B Instruct  & 0.86 & 0.82 & 0.80 & 0.83 & 0.84 \\ 
    Llama 3 70B & 0.86 & 0.57 & 0.70 & 0.58 & 0.71 \\
    Llama 3 8B Instruct & 0.64 & 0.75 & 0.70 & 0.67 & 0.70 \\
    Llama 2 70B & 0.68 & 0.71  & 0.65 & 0.67 & 0.70 \\
    Mixtral 7x8B Instruct & 0.75 & 0.64 & 0.70 & 0.67 & 0.70 \\
    Llama 2 70B Chat & 0.82 & 0.53 & 0.55 & 0.83 & 0.68 \\
    Llama 2 7B Chat & 0.89 & 0.39 & 0.55 & 0.75 & 0.64 \\ 
    Llama 2 7B & 0.57 & 0.64 & 0.45 & 0.67 & 0.61 \\ 
    Llama 3 8B & 0.89 & 0.25  & 0.40 & 0.75 & 0.57  \\
    Llama 2 13B Chat & 0.93 & 0.21 & 0.50 & 0.58 & 0.57 \\
    Llama 2 13B & 0.89 & 0.07  & 0.45 & 0.50 & 0.48  \\
    Random baseline & 0.4 & 0.4 & 0.4 & 0.4 & 0.4 \\
    \bottomrule
    \end{tabular}
    \caption{Accuracy on the zero-shot context-based inference task (Experiment 1c) overall, by bigram frequency and by context bias. 
    }
    \label{tab:exp-1c-context-0shot}
\end{table*}

\end{document}